\documentclass{article}

 \usepackage[main, final]{neurips_2026}

\usepackage[utf8]{inputenc} 
\usepackage[T1]{fontenc}    
\usepackage{hyperref}       
\usepackage{url}            
\usepackage{booktabs}       
\usepackage{amsfonts}       
\usepackage{nicefrac}       
\usepackage{microtype}      
\usepackage{xcolor}         

\usepackage{amsmath}
\usepackage{amssymb}
\usepackage{multirow}
\usepackage[table]{xcolor}
\usepackage{graphicx}
\usepackage{enumitem}

\usepackage{subcaption}
\usepackage{placeins}
\usepackage{float}
\newcommand{\partitle}[1]{\smallskip \noindent \textbf{#1.}}
\title{TSQAgent: Rating Time Series Data Quality via Dedicated Agentic Reasoning}

\author{
Shunyu Wu$^{1,2}$, \ Dan Li$^{1}$\thanks{Dan Li is the Corresponding Author.}, \ Haozheng Ye$^{1,2}$, \ Weibin Feng$^{1,2}$, \ Jian Lou$^{1,2}$, \\{\bfseries Bo Zhang$^{3}$, \ Wenjie Feng$^{4}$,  \ Chenjuan Guo$^{5}$, \ See-Kiong Ng$^{6}$} \\
$^{1}$Sun Yat-sen University \\
$^{2}$State Key Laboratory of Internet Architecture \quad
$^{3}$China University of Mining and Technology \\
$^{4}$University of Science and Technology of China \quad
$^{5}$East China Normal University \\
$^{6}$National University of Singapore \\
\texttt{wushy88@mail2.sysu.edu.cn}, \ \texttt{lidan263@mail.sysu.edu.cn}
}

\begin{document}

\maketitle

\begin{abstract}
Assessing the quality of time series (TS) data is fundamental yet inherently challenging due to the multifaceted nature of quality dimensions. Recently, large language models (LLMs) have emerged as a promising paradigm for TS quality assessment via pairwise comparison and per-dimension evaluation. However, existing approaches rely on manually predefined quality dimensions and purely text-based reasoning, leaving it unknown whether LLMs can identify truly relevant quality dimensions or perform grounded and quantitative quality comparisons. 
To investigate this, we construct TSQBench, a dedicated benchmark for evaluating LLMs on two progressive capabilities: (i) understanding and identifying relevant quality dimensions, and (ii) performing quality comparison under specific dimensions. Our analysis reveals that current LLMs consistently struggle with both dimension identification and evidence-grounded quality comparison. 
To address these limitations, we propose \textsc{TSQAgent}, a novel agentic reasoning framework for TS quality rating consisting of three collaborative roles: \emph{Perceiver} for focused dimension selection, \emph{Inspector} for dimension-wise quantitative analysis, and \emph{Adjudicator} that aggregates and refines the final judgment. In particular, we introduce an agentic reasoning strategy that instills the ability to identify and prioritize the most relevant quality dimensions, and further propose an agent workflow equipped with external analytical tools to enable precise quantitative comparisons over selected dimensions. 
Experiments on both the proposed benchmark and eleven real-world datasets demonstrate that our framework not only substantially improves LLMs' capabilities in quality understanding and quantitative comparison but also effectively translates these improvements into better quality-aware data selection, leading to enhanced downstream performance and data efficiency. Code is available at \url{https://github.com/clsr1008/TSQAgent}.
\end{abstract}

\section{Introduction}
\label{intro}
The quality of time series (TS) data plays a critical role in building performant TS models, as low-quality data can substantially degrade model capability.  In real-world scenarios, TS data is notoriously affected by various quality issues, including missing values, outliers, noise contamination, distribution shifts, and structural distortions (e.g., trend or frequency inconsistencies)~\citep{fang2024bayotide,zhang2024irregular,liu2024time}. These issues render quality assessment of TS data inherently difficult, as it requires synthesizing multiple interacting quality dimensions and accurately assessing each dimension according to its specific evaluation criteria.
Existing approaches primarily rely on statistical feature analysis~\citep{jadidi2023correlation} and heuristic rules~\citep{rahm2000data} to identify low-quality samples, or extend general data attribution (Influence Functions~\citep{hampel1974influence} or Shapley~\citep{shapley1953value}) to TS quality~\citep{zhang2025timeinf,bento2021timeshap}. A more detailed discussion of related approaches is provided in Section~\ref{rw}.

Recently, propelled by the success of large language models (LLMs) in a range of TS tasks from forecasting~\citep{liu2024lstprompt,gruver2023large} to TS question answering (TSQA)~\citep{kong2025time,guan2025timeomni,langer2025opentslm}, there has been growing interest in leveraging their intrinsic TS knowledge to enable comprehensive, multi-dimensional TS quality assessment.
Along this direction, recent works have explored the use of LLMs for time series understanding and quality-related analysis. For example, TimeSeriesExam~\citep{cai2024timeseriesexam} constructs a benchmark for evaluating LLMs' understanding of temporal patterns (e.g., trends and periodicity) and quality-related signals, while AXIS~\citep{lan2025axis} leverages LLMs to interpret anomalous patterns as potential indicators of data quality issues. These studies suggest that LLMs possess a certain level of temporal understanding and pave a promising foundation for time series quality assessment. Building upon this, TSRating~\citep{wu2026rating} explores LLM-based TS quality assessment by evaluating data along four predefined dimensions (namely trend, frequency, amplitude, and pattern).

\partitle{Gap and Research Question} While such approaches establish a structured and automated assessment pipeline, they still rely on manually specified quality criteria and predominantly text-based reasoning, leaving the underlying capabilities of LLMs for TS quality assessment largely unexplored. This motivates us to investigate the following research question: \textit{Can LLMs understand and adaptively identify relevant quality dimensions, and perform evidence-grounded quality comparisons under specific quality dimensions?}

\partitle{Our Work}
To investigate this question, we construct TSQBench, a dedicated benchmark to evaluate two capabilities of LLMs: quality dimension identification and evidence-grounded quality comparison.
Specifically, we design a comprehensive set of quality dimensions, and leverage controlled quality degradation guided by semantic factors to generate paired time series with diverse and fine-grained quality differences~\citep{xie2025chatts}. The benchmark results reveal two gaps in LLM-based TS quality assessment: (i) a lack of awareness of which quality dimensions are truly informative for selecting high-quality TS data~\citep{zhang2026many}, and (ii) an inherent weakness in reliable numerical reasoning over time series signals~\citep{wu2026timeart}, which often leads to unstable or hallucinated quality judgments.

Motivated by these findings, we propose \textsc{TSQAgent}, a novel agentic reasoning framework for TS quality assessment and pairwise comparison. The agent system decomposes the task into three collaborative roles: (1) \emph{Perceiver} builds a high-level understanding of the input pair and selects relevant quality dimensions for evaluation; (2) \emph{Inspector} performs dimension-wise analysis through iterative reasoning with action-based evidence acquisition in a ReAct-style manner~\citep{yao2023react}; and (3) \emph{Adjudicator} aggregates all evidence into the final judgment while performing reflection to ensure consistency and completeness~\citep{abbas2025pitch,li2025flow}.
In particular, we introduce an agentic reasoning strategy~\citep{shao2024deepseekmath} that leverages TSQBench to train the Perceiver via tailored Group Relative Policy Optimization (GRPO) for adaptive and focused quality dimension selection, moving beyond fixed predefined criteria toward structured reasoning over heterogeneous quality factors. Next, we design a tool-augmented agent workflow that equips the Inspector with external analytical tools for grounded quantitative comparison over selected dimensions, where tool outputs are explicitly injected into the reasoning process as evidence~\citep{wei2026agentic,zhang2026recthinker}. Our tool suite consists of 16 specific TS analytical functions covering statistical, spectral, and structural measurements.

We conduct experiments on both TSQBench and eleven real-world datasets, covering a full pipeline from pairwise quality judgments to training a dedicated quality rating model and downstream quality-aware data selection. The results demonstrate that our framework not only substantially improves LLMs’ capabilities in quality understanding and quantitative comparison but also effectively translates these improvements into better data selection performance, leading to enhanced downstream accuracy and data efficiency. Notably, we fine-tune Timer-S1~\citep{liu2026timer}, one of the largest publicly available TS foundation models with 8.3B parameters, and show that using only 75\% of the data selected by our method achieves performance comparable to training on the full data pool.

\partitle{Summary of Contributions} Our main contributions are summarized as follows:
\begin{itemize}[leftmargin=*]
    \item We identify and formalize the problem of TS quality assessment for LLMs, and construct a dedicated benchmark to evaluate two key capabilities of LLMs: TS quality dimension identification and evidence-grounded quality comparison.
    \item We propose \textsc{TSQAgent}, an agentic reasoning framework that integrates reasoning-driven dimension selection and tool-augmented quantitative analysis to address the core challenges of TS quality assessment.
    \item We conduct extensive experiments demonstrating that our approach improves both LLM capabilities and downstream data selection performance across diverse datasets and models.
\end{itemize}

\section{Measuring LLM Capabilities in Time Series Quality Rating}
\label{sec:benchmark}

To systematically evaluate LLMs in time series quality understanding, we construct a synthetic benchmark, named TSQBench, based on pairwise comparison. TSQBench is designed to assess two core capabilities: (i) \textbf{quality dimension identification}, i.e., whether an LLM can correctly identify which quality dimensions (e.g., noise level or frequency patterns) two TS segments differ on; and (ii) \textbf{quality comparison}, i.e., whether an LLM can correctly decide which series exhibits higher quality under a given dimension.

\partitle{Benchmark Design}
TSQBench is built on a controlled synthetic generation pipeline, where paired time series are derived from a shared template and then injected with dimension-specific degradations. We define a unified quality dimension set $\mathcal{D}=\{d_1,\ldots,d_7\}$, including missing value, noise level, rare pattern, trend, frequency, amplitude, and pattern consistency, as detailed in Table~\ref{tab:quality_dimension}. These dimensions describe data integrity, signal irregularities, and structural properties of temporal dynamics, providing a comprehensive basis for quality comparison~\citep{cleveland1990stl,hyndman2018forecasting,kazemi2019time2vec}. Each benchmark instance activates a subset of dimensions, where asymmetric perturbations are applied to one side, enabling controlled and interpretable quality differences. Further details on TSQBench construction are provided in Appendix~\ref{app:benchmark}.

\partitle{Evaluation and Findings}
We evaluate different LLMs on TSQBench using 1,000 synthetic samples, where Precision/Recall are used for dimension identification and accuracy is used for quality comparison.
The results in Table~\ref{tab:llm_tsqa} reveal two consistent limitations of existing LLMs in TS quality reasoning. For dimension identification, models exhibit low precision (33.4\%--43.0\%) despite relatively high recall (63.9\%--83.0\%), indicating unstable dimension selection behavior and a tendency to over-select irrelevant quality dimensions. For quality comparison, all models achieve only marginal improvements over random guessing (50\%), with accuracies ranging from 54.7\% to 61.0\%, suggesting limited intrinsic capability in performing reliable fine-grained quality comparison. These observations motivate the two core designs in TSQAgent proposed in Section~\ref{sec:agent}: (i) a reasoning-oriented dimension selection mechanism trained via TS quality-tailored GRPO, and (ii) a tool-augmented agentic reasoning for quantitative quality comparison.

\begin{table}[!htbp]
  \vskip -1.0em
  \caption{Performance of LLM-based TS quality assessment on TSQBench. We evaluate two core capabilities: dimension identification and quality comparison.}
  \label{tab:llm_tsqa}
  \centering
  \small
  \setlength{\tabcolsep}{4pt}
  \begin{tabular}{l | c c c c c c}
    \toprule
    \textbf{Model} & Qwen3-4B & Gemma4-E4B & Phi4-mini & Qwen3-8B & GPT-4o mini & Claude Haiku 4.5 \\
    \midrule
    \multicolumn{7}{c}{\underline{~~~Dimension Identification~~~}} \\
    Precision & 33.5\% & 38.0\% & 33.4\% & 37.4\% & 34.7\% & 43.0\% \\
    Recall    & 80.9\% & 75.7\% & 69.7\% & 78.0\% & 83.0\% & 63.9\% \\
    \midrule
    \multicolumn{7}{c}{\underline{~~~Quality Comparison~~~}} \\
    Accuracy  & 54.7\% & 58.6\% & 54.9\% & 58.4\% & 60.0\% & 61.0\% \\
    \bottomrule
  \end{tabular}
\end{table}

\section{TSQAgent: Agentic Reasoning Framework Tailored for TS Quality}
\label{sec:agent}

\begin{figure*}[t]
    \centering
    \includegraphics[width=1.0\textwidth]{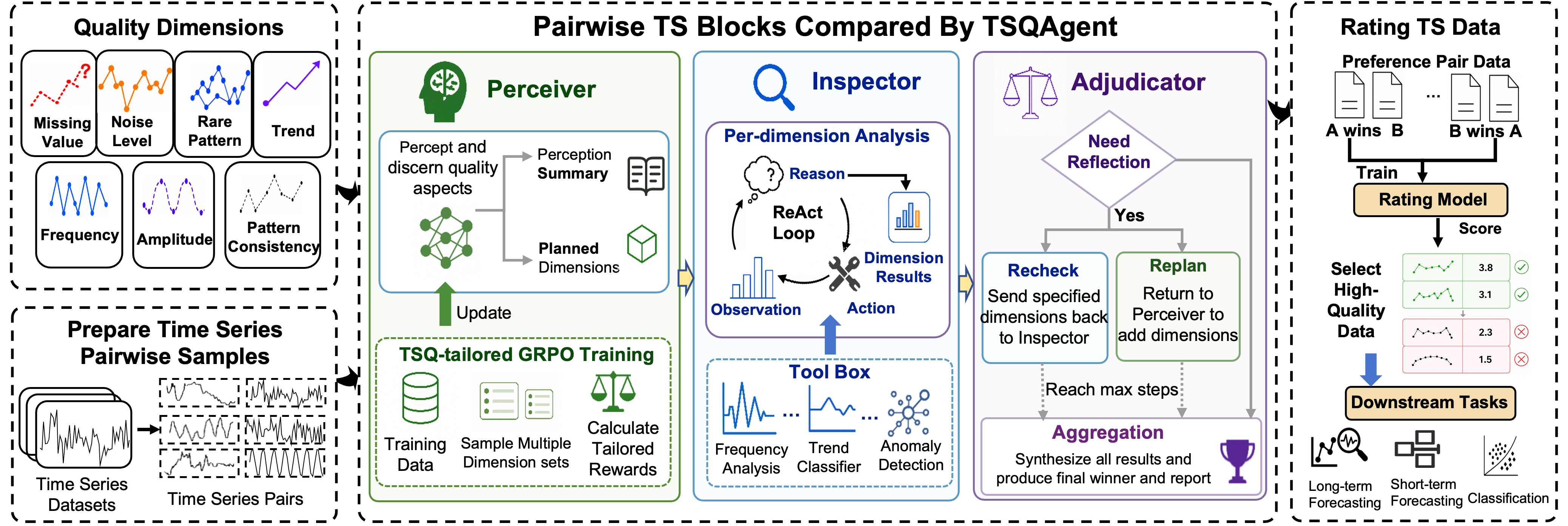}
    \vskip -0.5em
    \caption{Overview of the proposed TSQAgent framework for time series quality rating.}
    \label{fig:framework}
\end{figure*}

\subsection{Overview of TSQAgent}

TSQAgent is a multi-agent framework for pairwise time series quality assessment, designed to determine which of two time series exhibits higher overall quality through structured reasoning and tool-grounded analysis. The pairwise formulation simplifies evaluation via relative comparison, avoiding absolute scoring and improving robustness across diverse scenarios~\citep{peng2025agentic}. The system decomposes the task into three specialized agents that operate collaboratively: (1) \textbf{Perceiver}: constructs a high-level understanding of the input pair and selects relevant quality dimensions for evaluation; (2) \textbf{Inspector}: performs dimension-wise analysis via iterative reasoning and tool-based evidence acquisition; and (3) \textbf{Adjudicator}: aggregates dimension-level evidence into a final judgment while performing reflection to ensure consistency and completeness. The overall architecture of TSQAgent is shown in Figure~\ref{fig:framework}.

Formally, given two time series segments $(\mathrm{series}_A, \mathrm{series}_B)$ and dataset context $c$, the framework produces a final comparative result $R = (w, s, C)$, where $w$ denotes the higher-quality series, $s$ the confidence level, and $C$ a structured explanation grounded in multi-dimensional evidence. 
Specifically, the confidence level $s \in [0,1]$ measures the reliability of each pairwise judgment based on the consistency and sufficiency of the available evidence.

To complement the data rating and selection process, we convert pairwise preferences into scalar quality scores. Specifically, TSQAgent first produces preference judgments between time series pairs. We then filter out low-confidence comparisons and train a lightweight rating model on these comparisons to assign a scalar quality score to each sample via Bradley--Terry optimization, thereby enabling global ranking and quality-aware data selection.
The rating model architecture and training details are provided in Appendix~\ref{app:rating_model}.

\subsection{Perceiver: Perception and Planning}

The Perceiver serves as the entry point of TSQAgent, responsible for constructing a high-level understanding of the input pair $(\mathrm{series}_A, \mathrm{series}_B)$ and selecting relevant quality dimensions for subsequent analysis. Formally, the Perceiver can be viewed as a mapping
\begin{equation}
(Z, \mathcal{D}_{\text{plan}}) = \mathcal{A}_p(c, \mathrm{series}_A, \mathrm{series}_B),
\end{equation}
where $c$ denotes the dataset description, and the output consists of a perception summary $Z$, and a subset of selected quality dimensions $\mathcal{D}_{\text{plan}}$ for subsequent inspection.

Given the raw time series inputs and associated statistical summaries (e.g., mean, standard deviation, quantiles), the Perceiver constructs a perception summary $Z$ that captures the salient differences between the two series. This reasoning process is guided by a predefined set of quality dimensions $\mathcal{D} = \{d_1, \ldots, d_7\}$ (see Table~\ref{tab:quality_dimension}), which characterizes key quality aspects such as data completeness, noise characteristics, and temporal structure. Based on this, the Perceiver selects a subset of relevant dimensions $\mathcal{D}_{\text{plan}} \subseteq \mathcal{D}$, emphasizing the most informative aspects for subsequent analysis.

\partitle{Agentic Reasoning for TS Quality Dimension Selection} 
Importantly, we formulate time series quality dimension selection as an agentic reasoning process within the Perceiver, where the agent is required to infer and prioritize the most decision-relevant quality aspects for each input pair. To instill this capability, we introduce a TS quality-tailored GRPO (Group Relative Policy Optimization) training strategy~\citep{shao2024deepseekmath}, which serves as the underlying optimization framework. Our training dataset consists of 3,320 samples derived from TSQBench pipeline (see Section~\ref{sec:benchmark}), and the optimization is guided by a structured reward that enforces both output validity and alignment between predicted and ground-truth dimension sets. This induces more discriminative reasoning over time series quality dimensions, improving fine-grained differentiation across diverse series pairs. More implementation details including training configurations are provided in Appendix~\ref{app:grpo}.

\subsection{Inspector: Dimension-wise Analysis}
\label{sec:inspect}

The Inspector is responsible for fine-grained evaluation of each selected quality dimension and serves as the core reasoning module in TSQAgent. Given the planned dimensions $\mathcal{D}_{\text{plan}}$ and perception summary $Z$, it performs pairwise comparison between both series, producing structured judgments for each dimension.

The Inspector executes evaluation in a dimension-by-dimension manner over $\mathcal{D}_{\text{plan}}$, performing pairwise assessment between $\mathrm{series}_A$ and $\mathrm{series}_B$ under a bounded ReAct-style interaction~\citep{yao2023react}.
Formally, for each dimension $d$, the process proceeds through an iterative reasoning-action loop:
\begin{equation}
q_d \rightarrow (\text{Reason}_t \rightarrow \text{Action}_t \rightarrow \text{Observation}_t)^{*_{K}} \rightarrow R_d.
\end{equation}
Here, $q_d$ denotes the evaluation query derived from the input pair $(\mathrm{series}_A, \mathrm{series}_B)$ and the specification of dimension $d$. Each iteration consists of three components: $\text{Reason}_t$ performs internal reasoning over observed signals and evidence; $\text{Action}_t$ optionally invokes external analytical tools to obtain quantitative evidence; and $\text{Observation}_t$ captures the corresponding tool outputs or retrieved signals.
$R_d = (w_d, s_d, E_d, C_d)$ represents the final assessment, including the predicted winner $w_d \in \{A, B, \text{TIE}\}$, confidence score $s_d \in [0,1]$, supporting evidence $E_d$, and textual conclusion $C_d$.
$K$ denotes the maximum number of reasoning-action iterations per dimension, ensuring that each dimension is evaluated within a limited computational budget, after which the Inspector finalizes the outputs based on accumulated evidence.

\partitle{Tool-Augmented Quantitative Quality Comparison}
Within each ReAct loop, the Inspector performs dimension-specific reasoning over raw time series, leveraging internal knowledge and accumulated evidence to identify salient patterns relevant to the target dimension, such as trend shifts, periodic structures, or rare behaviors. When internal reasoning is insufficient to resolve subtle differences, the Inspector invokes external analytical tools to obtain quantitative evidence~\citep{qian2025smart,schick2023toolformer}. In contrast to purely qualitative reasoning, tool-based signals provide measurable and fine-grained evidence that is often difficult to infer directly from raw sequences, thereby improving decision robustness~\citep{wu2026timeart}.
To support this process, we develop a library of analytical tools for data completeness, noise, rare pattern detection, and temporal structure analysis, as detailed in Appendix~\ref{app:tools}. Tool outputs are treated as structured observations within the reasoning trajectory, and are cached for reuse across dimensions to reduce redundant computation.

\subsection{Adjudicator: Reflection and Aggregation}

The Adjudicator is responsible for producing a final comparative judgment between the two time series by aggregating dimension-level assessment results from the Inspector, while also ensuring the completeness and consistency of the evidence through a reflection mechanism.

Formally, given the dimension-level results $\mathcal{R} = \{R_d \mid d \in \mathcal{D}_{\text{plan}}\}$ with $R_d = (w_d, s_d, E_d, C_d)$ produced by the Inspector, the Adjudicator first performs a reflection step over $\mathcal{R}$ to evaluate whether the available evidence is sufficient and mutually consistent. This process yields a reflection directive $\rho \in \{\texttt{recheck}(\mathcal{D}_{\text{recheck}}), \texttt{replan}, \varnothing\}$, where $\varnothing$ indicates that no further refinement is required.

The reflection directive distinguishes two types of refinement. A $\texttt{recheck}(\mathcal{D}_{\text{recheck}})$ signal is generated when the evidence for certain dimensions is insufficient or internally conflicting, where $\mathcal{D}_{\text{recheck}} \subseteq \mathcal{D}_{\text{plan}}$ specifies the subset of dimensions to be re-evaluated by the Inspector. Alternatively, a $\texttt{replan}$ signal is issued when the current dimension set $\mathcal{D}_{\text{plan}}$ is incomplete with respect to important aspects of data quality, requiring the Perceiver to introduce additional dimensions for further assessment. These reflection signals enable iterative refinement of the evaluation process across agents~\citep{shinn2023reflexion}. To ensure stable execution, reflection steps are explicitly bounded. When the budget is exhausted, the reflection process terminates and is treated as $\rho = \varnothing$, after which the Adjudicator proceeds to aggregate all dimension-level results, thereby preventing indefinite refinement loops.

When the evidence is considered sufficient and consistent ($\rho = \varnothing$), the Adjudicator aggregates the dimension-level results by considering the predicted winners, confidence scores, supporting evidence, and textual conclusions. It then produces the final judgment $R = (w, s, C)$.
Further analysis of dimension-level interpretability and task-dependent quality assessment is provided in Appendix~\ref{app:dimension_interpretability}.

\section{Experiments}
\label{sec:exp}

We evaluate TSQAgent from two complementary perspectives:
(1) controlled evaluation on TSQBench to verify how reasoning-driven dimension selection and tool augmentation improve the LLMs' capabilities of TS quality evaluation;
(2) real-world data selection experiments to evaluate the effectiveness of TSQAgent across diverse datasets and downstream models. 
Additional qualitative case studies are provided in Appendix~\ref{app:case_study} to illustrate the reasoning behaviors of the proposed agents.

\subsection{Capability Evaluation on TSQBench}

We conduct capability evaluation on TSQBench to separately assess the two core capabilities of LLMs: dimension identification and dimension-wise quality comparison. We evaluate a diverse set of both open-source and proprietary models, including Qwen3~\citep{yang2025qwen3}, Gemma4~\citep{Kamath2025Gemma3T}, Phi4~\citep{abdin2024phi}, GPT-4o mini~\citep{achiam2023gpt}, and Claude Haiku 4.5. To examine whether the capabilities learned from TSQBench generalize beyond synthetic quality degradations, we provide an additional evaluation on real-world anomalies in Appendix~\ref{app:realworld_generalization}.

\begin{figure}[!htbp]
    \centering
    \begin{subfigure}[t]{0.42\linewidth}
        \centering
        \includegraphics[width=\linewidth]{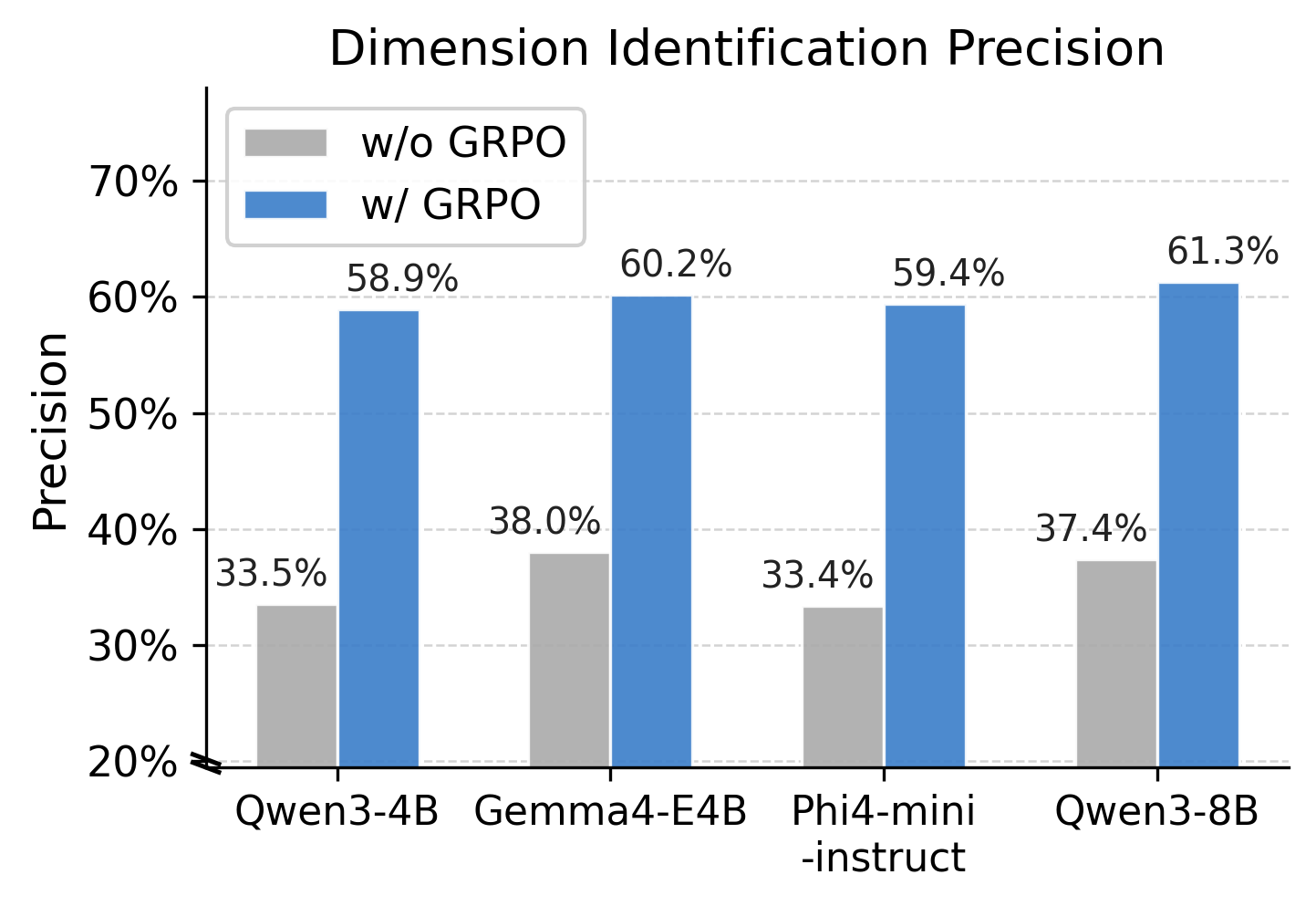}
        \vskip -0.5em
        \caption{Effect of GRPO on Dimension Identification}
        \label{fig:unit_dim_id}
    \end{subfigure}
    \hfill
    \begin{subfigure}[t]{0.57\linewidth}
        \centering
        \includegraphics[width=\linewidth]{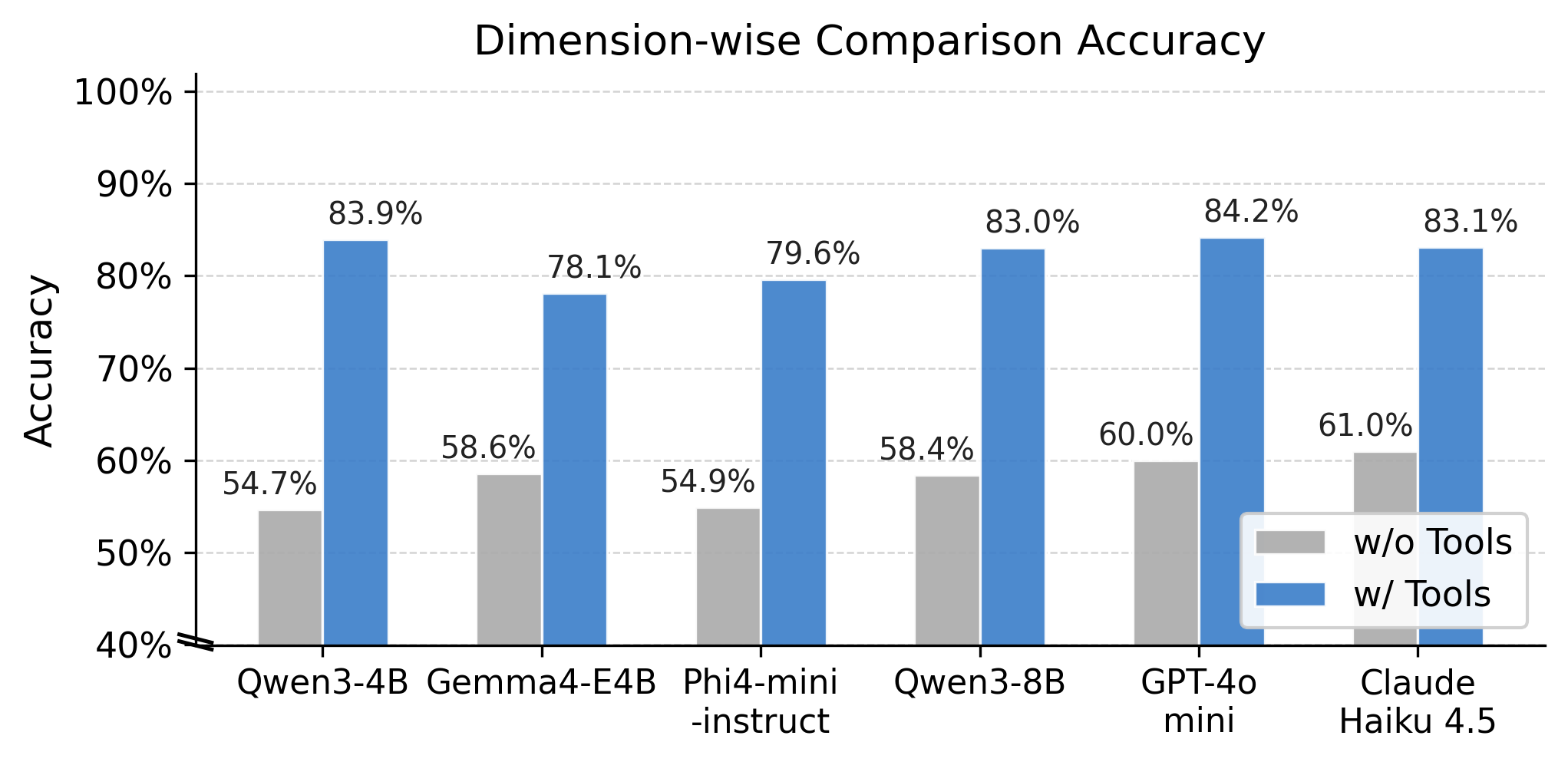}
        \vskip -0.5em
        \caption{Effect of Tool Augmentation on Quality Comparison}
        \label{fig:unit_quant_comp}
    \end{subfigure}
    \vskip -0.5em
    \caption{Capability Evaluation on TSQBench. 
    Subplot (a) reports dimension identification performance under different models and training configurations, highlighting the effect of reasoning-driven dimension selection. 
    Subplot (b) reports quantitative comparison performance with and without tool usage, showing the impact of external analytical support on fine-grained quality assessment.}
    \label{fig:unit_tests}
\end{figure}

\partitle{Dimension Identification Precision}
As shown in Figure~\ref{fig:unit_dim_id}, we evaluate the effect of our TS quality-tailored GRPO training strategy on dimension selection performance. The strategy consistently improves dimension identification precision across all evaluated open-source models, increasing precision from around 33\%--38\% to nearly 60\%. These results indicate that vanilla LLMs tend to over-select irrelevant quality dimensions, while our reasoning-based training strategy effectively encourages more focused and discriminative dimension selection behavior. Additional results using the F1 score are provided in Appendix~\ref{app:perceiver_f1}.

\partitle{Quantitative Comparison Accuracy}
Figure~\ref{fig:unit_quant_comp} reports the performance of models with and without tool access for quality comparison. Tool augmentation consistently improves comparison accuracy across all evaluated models, raising performance from approximately 55\%--61\% to over 78\%--84\%. The improvements are observed for both open-source and proprietary models, suggesting that external analytical tools provide critical quantitative evidence that is difficult to obtain through purely text-based reasoning alone.

\subsection{Real-World Data Selection Experiments}
\label{subsec:select}

The goal of this section is to evaluate whether TSQAgent can reliably assess data quality in real-world settings, and whether such assessments translate into improved downstream performance and data efficiency. By selecting high-quality subsets based on the TSQAgent’s judgments, we test whether comparable or superior performance can be achieved with reduced training data, thereby complementing the results on the synthetic TSQBench.

\partitle{Datasets and Baselines}
We conduct experiments on 11 widely-used real-world datasets across three task categories. For long-term forecasting, we use \textit{Electricity}~\citep{trindade2015electricityloaddiagrams20112014}, \textit{ExchangeRate}~\citep{lai2018modeling}, \textit{Traffic}~\citep{lai2018modeling}, and \textit{Weather}~\citep{wu2021autoformer}. For short-term forecasting, we adopt the \textit{M4 dataset}~\citep{makridakis2018m4}, including its \textit{Yearly}, \textit{Monthly}, and \textit{Daily} subsets. For classification, we use four datasets from the UCR/UEA archive~\citep{ismail2019deep}: \textit{MedicalImages}, \textit{CBF}, \textit{BME}, and \textit{Handwriting}.
We compare our method with several representative data valuation and selection baselines, including DataShapley~\citep{ghorbani2019data}, DataOob~\citep{kwon2023data}, TimeInf~\citep{zhang2025timeinf}, and TSRating~\citep{wu2026rating}. All baselines produce sample-level scores. Details of the datasets and baselines are provided in the Appendix~\ref{app:configs}.

\partitle{Setups and Evaluation}
For all tasks, we construct training, validation, and test splits with a ratio of 7:2:1 and then segment time series into fixed-length samples according to the requirements of each dataset (e.g., historical context length determined by the forecasting configuration). Each resulting sample is assigned a quality score by different methods, and we select the top 50\% of samples to train downstream models (e.g., CNN, PatchTST~\citep{nie2023time}). The trained models are then evaluated on the held-out test set under standard metrics. We adopt task-specific metrics: Root Mean Squared Error (RMSE) for long-term forecasting, Mean Absolute Percentage Error (MAPE) for short-term forecasting, and Accuracy for classification. Better data selection is reflected by improved downstream performance. All results are averaged over 10 runs with different random seeds.

\begin{table}[!htbp]
  \caption{Comparison of data selection methods under a 50\% high-quality data budget across forecasting and classification tasks on multiple datasets and models. Best results are \textbf{bolded}.}
  \label{tab:all_tasks}
  \centering
  \footnotesize
  \setlength{\tabcolsep}{2.8pt}
  \begin{tabular}{l l | c c c c | c c c | c c c c}
    \toprule
    \textbf{Model} & \textbf{Method} &
    \multicolumn{4}{c|}{\textbf{Long-term (RMSE $\downarrow$)}} &
    \multicolumn{3}{c|}{\textbf{Short-term (MAPE $\downarrow$)}} &
    \multicolumn{4}{c}{\textbf{Classification (Accuracy $\uparrow$)}} \\
    & &
    Elec. & ExRate & Traffic & Wea. &
    M4Y & M4M & M4D &
    MImg & CBF & BME & HW \\
    \midrule
    \multirow{5}{*}{Linear}
        & Random      & 1.601 & 0.356 & 0.979 & 0.665 & 1.705 & 1.208 & 1.672 & 0.390 & 0.294 & 0.427 & 0.053 \\
        & DataOob     & 1.539 & 0.318 & 0.761 & 0.638 & 1.949 & 1.133 & 1.779 & 0.457 & 0.318 & 0.260 & 0.028 \\
        & DataShapley & 1.580 & 0.323 & 0.956 & 0.638 & \textbf{1.488} & 1.207 & 1.370 & 0.432 & 0.337 & 0.433 & 0.038 \\
        & TimeInf     & 1.391 & 0.272 & \textbf{0.609} & 0.616 & 1.966 & 1.178 & 1.463 & 0.428 & 0.320 & 0.500 & 0.042 \\
        & TSRating    & 1.390 & 0.275 & 0.683 & 0.611 & 1.577 & 1.112 & 1.409 & 0.459 & 0.361 & 0.507 & 0.049 \\
        \rowcolor{cyan!10}
        & \textbf{TSQAgent} & \textbf{1.273} & \textbf{0.270} & 0.611 & \textbf{0.593} & 1.494 & \textbf{1.072} & \textbf{1.352} & \textbf{0.488} & \textbf{0.379} & \textbf{0.520} & \textbf{0.056} \\
    \midrule
    \multirow{5}{*}{CNN}
        & Random      & 1.592 & 1.474 & 0.504 & 0.769 & 2.332 & 1.124 & 1.193 & 0.554 & 0.595 & 0.495 & 0.151 \\
        & DataOob     & 1.609 & 1.527 & 0.552 & 0.737 & 2.957 & 1.208 & 1.346 & 0.514 & 0.663 & 0.521 & 0.155 \\
        & DataShapley & 1.529 & 1.598 & 0.475 & 0.767 & 2.553 & 1.117 & 1.159 & 0.550 & 0.593 & 0.564 & 0.159 \\
        & TimeInf     & 1.530 & 1.515 & 0.505 & 0.758 & 2.289 & 1.117 & 1.103 & 0.567 & 0.563 & 0.535 & 0.155 \\
        & TSRating    & 1.511 & 1.429 & 0.428 & 0.734 & 1.782 & \textbf{1.075} & 1.108 & 0.561 & \textbf{0.679} & 0.575 & 0.159 \\
        \rowcolor{cyan!10}
        & \textbf{TSQAgent} & \textbf{1.419} & \textbf{1.371} & \textbf{0.412} & \textbf{0.723} & \textbf{1.744} & 1.108 & \textbf{1.096} & \textbf{0.574} & 0.668 & \textbf{0.580} & \textbf{0.162} \\
    \midrule
    \multirow{5}{*}{PatchTST}
        & Random      & 0.406 & 0.222 & 0.361 & 0.474 & 4.146 & 2.044 & 1.985 & 0.561 & 0.470 & 0.475 & 0.124 \\
        & DataOob     & 0.416 & 0.226 & 0.362 & 0.558 & 3.982 & 2.060 & 1.958 & 0.571 & 0.489 & 0.489 & 0.141 \\
        & DataShapley & \textbf{0.396} & \textbf{0.212} & 0.361 & 0.457 & 4.273 & 2.017 & 2.028 & 0.571 & 0.472 & 0.529 & 0.126 \\
        & TimeInf     & 0.415 & 0.220 & 0.351 & 0.510 & 4.029 & 2.019 & 2.023 & 0.522 & 0.467 & 0.504 & 0.131 \\
        & TSRating    & 0.397 & 0.213 & 0.351 & 0.467 & 3.901 & 1.957 & 1.863 & 0.572 & 0.511 & \textbf{0.536} & 0.156 \\
        \rowcolor{cyan!10}
        & \textbf{TSQAgent} & 0.405 & 0.214 & \textbf{0.348} & \textbf{0.450} & \textbf{3.820} & \textbf{1.856} & \textbf{1.855} & \textbf{0.573} & \textbf{0.516} & 0.533 & \textbf{0.166} \\
    \bottomrule
  \end{tabular}
\end{table}

\subsubsection{Main Results}
\label{subsec:mainresults}
Table~\ref{tab:all_tasks} presents the performance of different data selection methods across three downstream time series tasks with three representative downstream model architectures: Linear, CNN-based, and Transformer-based (PatchTST). Overall, our method consistently achieves the best or near-best performance across most settings, demonstrating strong effectiveness in identifying high-quality data. In long-term forecasting, it attains the lowest RMSE in the majority of cases, particularly showing clear advantages under Linear and CNN models. For short-term forecasting, our method obtains the best MAPE in 7 out of 9 cases, indicating robust performance across different temporal granularities. Likewise, in classification tasks, it ranks first in 10 out of 12 cases, suggesting that the selected subsets effectively preserve discriminative information. Compared with TSRating and TimeInf, our method shows more consistent gains across settings, indicating more accurate and effective quality assessment for guiding data selection. The results with standard deviations across random seeds are reported in Appendix~\ref{app:std}.

\subsubsection{Component Analysis}
\label{subsec:ablation}
\begin{figure}[!htbp]
    \centering
    \includegraphics[width=0.9\linewidth]{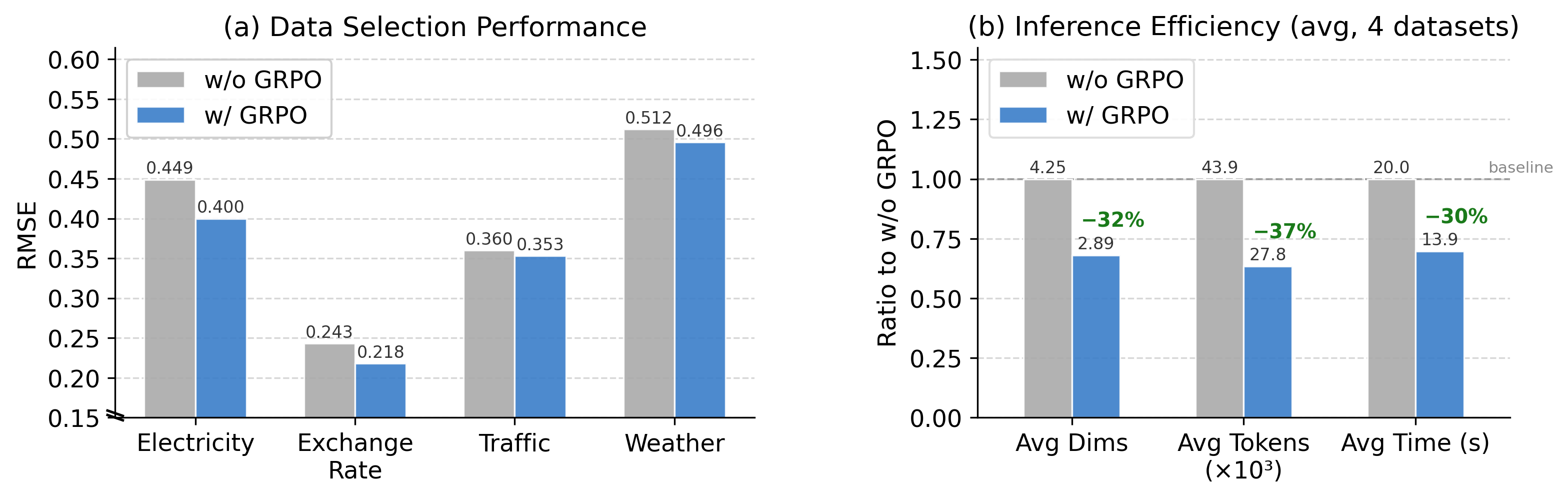}
    \vskip -0.5em
    \caption{Effect of reasoning-driven Perceiver training. The left figure shows downstream data selection performance (RMSE) across four datasets using PatchTST as the downstream model, while the right figure reports efficiency metrics averaged over four datasets, including selected dimensions, token usage, and inference time.}
    \label{fig:grpo_ablation}
\end{figure}

\partitle{Effect of Reasoning-driven Dimension Selection}
We evaluate the impact of TS quality-tailored GRPO on Perceiver, with results summarized in Figure~\ref{fig:grpo_ablation}. The left plot reports downstream long-term forecasting performance (RMSE) across four datasets using PatchTST, while the right plot shows the averaged efficiency metrics, including selected dimensions, token usage, and inference time.
Overall, our training strategy consistently improves data selection quality, leading to lower RMSE across all datasets. Meanwhile, it significantly reduces the number of selected dimensions as well as computational overhead in terms of token usage and inference latency. This suggests that the training strategy helps Perceiver focus on more informative dimensions while suppressing spurious or redundant reasoning paths, resulting in both more accurate and more efficient quality assessment.

\begin{figure}[!htbp]
    \centering
    \includegraphics[width=0.95\linewidth]{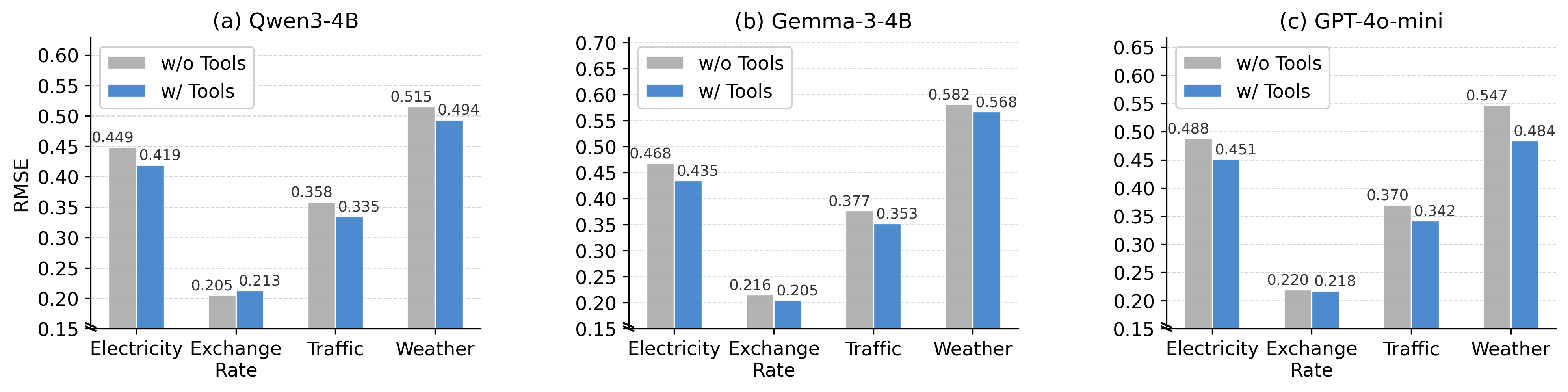}
    \vskip -0.5em
    \caption{Effect of tool-augmented quantitative comparison across different LLMs, where each subplot corresponds to a specific LLM backbone. The figures report downstream data selection performance (RMSE) under two settings: with tool usage (Inspector performs explicit quantitative comparison) and without tool usage (pure text-based reasoning). Results are evaluated on four datasets using PatchTST as the downstream model.}
    \label{fig:tool_ablation}
\end{figure}

\partitle{Effect of Tool-Augmented Quantitative Comparison}
We further study the impact of tool augmentation in the Inspector module, which enables explicit quantitative comparison, versus pure LLM-based reasoning without tool access. As shown in Figure~\ref{fig:tool_ablation}, we evaluate three different LLM backbones, i.e., Qwen3-4B~\citep{yang2025qwen3}, Gemma-3-4B~\citep{Kamath2025Gemma3T}, and GPT-4o mini~\citep{achiam2023gpt}, as the backbone of the TSQAgent across four datasets. The Inspector agent is required to perform data quality assessment under both settings, while downstream performance is measured by RMSE using PatchTST. The results indicate that enabling tool-based quantitative comparison generally improves performance across the majority of LLMs and datasets, suggesting that explicit quantitative signals enhance the reliability of data quality judgments and are broadly effective across different LLM backbones.

\begin{figure}[!htbp]
    \centering
    \includegraphics[width=0.95\linewidth]{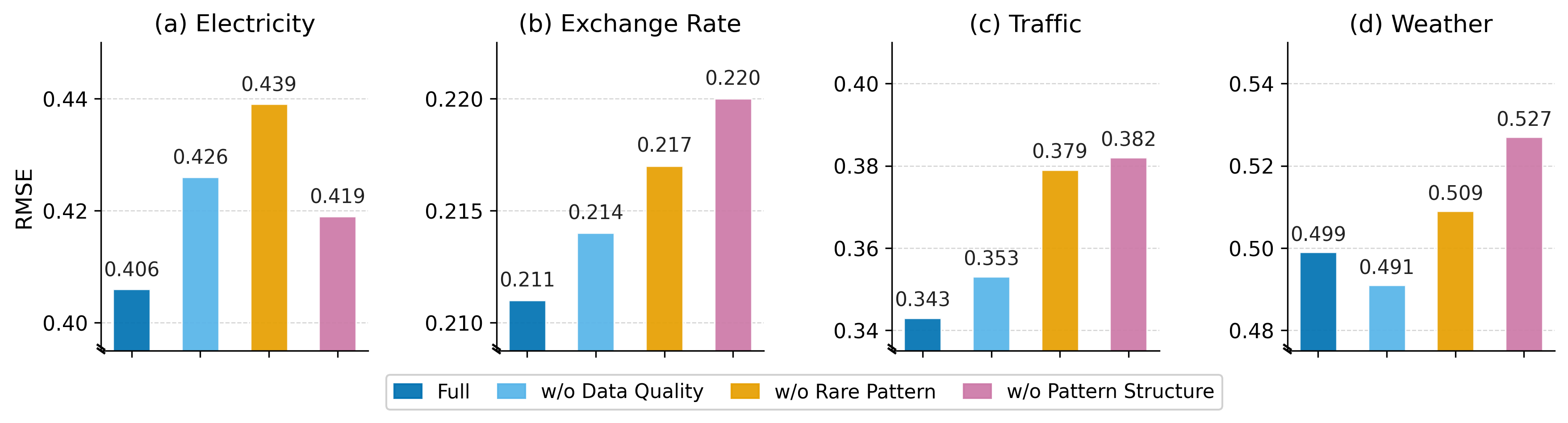}
    \vskip -0.5em
    \caption{Effect of quality dimension design. Each subplot shows results on a specific dataset, comparing downstream data selection performance (RMSE) under different dimension configurations, including the full set of seven dimensions and three ablated variants with one dimension group removed (Data Quality, Rare Pattern, or Pattern Structure). Results are evaluated on four datasets using PatchTST as the downstream model.}
    \label{fig:dimension_ablation}
\end{figure}

\partitle{Effect of Quality Dimension Design}
We further study the impact of quality dimension design as summarized in Table~\ref{tab:quality_dimension}. Specifically, we group the seven dimensions into three categories: \emph{Data Quality} (Missing Value, Noise Level), \emph{Rare Pattern} (Rare Pattern), and \emph{Pattern Structure} (Trend, Frequency, Amplitude, Pattern Consistency). We construct three ablated variants by removing each group in turn and compare them with the full configuration.
As shown in Figure~\ref{fig:dimension_ablation}, removing any group consistently leads to performance degradation in most cases. In particular, excluding \emph{Pattern Structure} or \emph{Rare Pattern} results in the largest drops, indicating the importance of capturing temporal structures and informative irregularities. Removing \emph{Data Quality} has a relatively smaller impact since such datasets exhibit limited variation in basic defects such as missing values and noise, making such signals less discriminative. Nevertheless, the full configuration achieves the best average performance, indicating that these dimension groups contribute complementary signals for reliable TS data quality assessment. The results with standard deviations for the component analysis experiments are reported in Appendix~\ref{app:std}.

\begin{table}[!htbp]
  \vspace{-1em}
  \caption{Cross-dataset data selection for TSFM (Timer-S1) fine-tuning. We compare random selection and quality-based selection under fixed data budgets, along with full-data training. Results are reported in RMSE (lower is better).}
  \label{tab:cross_dataset_tsfm}
  \centering
  \small
  \setlength{\tabcolsep}{4pt}
  \begin{tabular}{l | c c c c | c}
    \toprule
    \textbf{Config} & \textbf{Electricity} & \textbf{ExchangeRate} & \textbf{Traffic} & \textbf{Weather} & \textbf{Avg} \\
    \midrule
    Untrained & 0.477 & 0.228 & 0.293 & 0.510 & 0.377 \\
    \midrule
    Random (50\% data) & 0.448 & 0.225 & 0.290 & 0.492 & 0.364 \\
    \rowcolor{cyan!10}
    TSQAgent (50\% data) & 0.425 & 0.221 & 0.283 & 0.484 & 0.353 \\
    \midrule
    Random (75\% data) & 0.421 & 0.222 & 0.285 & 0.470 & 0.350 \\
    \rowcolor{cyan!10}
    TSQAgent (75\% data) & 0.400 & 0.218 & 0.278 & 0.468 & 0.341 \\
    \midrule
    Full (100\% data) & 0.395 & 0.218 & 0.274 & 0.465 & 0.338 \\
    \bottomrule
  \end{tabular}
\end{table}

\subsubsection{Cross-Dataset Selection for TSFM Fine-tuning}
\label{subsec:tsfm}

We further evaluate our method in a more challenging setting where data from multiple datasets are mixed for time series foundation model (TSFM) fine-tuning. The goal is to examine whether quality-aware selection remains effective under heterogeneous data distributions, and whether it can improve fine-tuning efficiency with limited data.
Specifically, we construct a unified pool by merging samples from four long-term forecasting training sets. To ensure cross-dataset comparability, we train the rating model via meta-learning to produce globally consistent quality scores, with details provided in Appendix~\ref{app:meta_learning}. We further examine how meta-training diversity affects the generalization of the rating model in Appendix~\ref{app:meta_generalization}. Based on these scores, we select the top 50\% of samples in the pool for fine-tuning Timer-S1~\citep{liu2026timer}, one of the largest publicly available TS foundation models, with 8.3B total parameters. The model is then evaluated on the test set of each dataset separately using RMSE. We compare against random selection with the same budget and full-data training. Additional comparisons with TSRating and DataShapley under the 50\% data budget are provided in Appendix~\ref{app:cross_dataset_selection}.

Table~\ref{tab:cross_dataset_tsfm} shows that our method consistently outperforms random selection across all datasets under both 50\% and 75\% data budgets, achieving lower RMSE with the same amount of training data. More importantly, when increasing the data budget to 75\%, the performance of our method becomes very close to full-data training (0.341 vs. 0.338 on average). This indicates that our approach can effectively identify high-quality training samples, enabling near-full-data performance with reduced data usage, thereby improving the efficiency of TSFM fine-tuning.

\subsubsection{Computational Cost and Efficiency}
\label{subsec:cost}

We further analyze the computational and API costs of TSQAgent under the cross-dataset TSFM selection setting described in Section~\ref{subsec:tsfm}. The unified training pool consists of four forecasting datasets (Electricity, ExchangeRate, Traffic, and Weather), containing 10,692 training samples in total. We collect 2,000 pairwise quality judgments through TSQAgent to train the shared meta rater.

\partitle{Computational Cost}
Table~\ref{tab:runtime_cost} reports the wall-clock time and amortized time per sample of TSQAgent and traditional data valuation baselines. TSQAgent achieves a favorable efficiency trade-off compared with traditional data valuation baselines, with an overall computational cost comparable to or lower than the comparison methods. Besides, the majority of TSQAgent's computational cost comes from the offline annotation stage, where the agentic framework is used to collect pairwise quality judgments. These judgments are then used to meta-train a shared rater across multiple datasets, enabling transferable quality assessment. Therefore, this upfront annotation and meta-training cost is a one-time expense, since the resulting meta rater can be efficiently adapted and reused.

At the inference stage, the trained rater model can be directly applied to new samples or datasets with only lightweight few-shot adaptation and rater inference costs. This stage requires only 64s in total (0.006s/sample), enabling efficient scaling to additional data. In comparison, although traditional data valuation methods do not require LLM-based annotation, they need to repeatedly compute sample importance for newly encountered samples or datasets, resulting in much higher amortized costs (e.g., 3.66s/sample for DataOob).

\begin{table}[!htbp]
  \vspace{-0.5em}
  \centering
  \caption{Computational cost comparison under the cross-dataset TSFM fine-tuning setting.}
  \label{tab:runtime_cost}
  \small
  \setlength{\tabcolsep}{6pt}
  \begin{tabular}{l|cc}
    \toprule
    \textbf{Method / Component} & \textbf{Wall-clock Time (s)} & \textbf{Amortized Time (s/sample)} \\
    \midrule
    DataShapley & 94197 & 8.81 \\
    DataOob & 39101 & 3.66 \\
    TimeInf & 28857 & 2.70 \\
    \rowcolor{cyan!10}
    \textbf{TSQAgent (Total)} & \textbf{24548} & \textbf{2.30} \\
    \hspace{1em}-- Pairwise judgment annotation & 24350 & 2.28 \\
    \hspace{1em}-- Meta-training & 134 & 0.013 \\
    \hspace{1em}-- Few-shot adaptation + inference & 64 & 0.006 \\
    \bottomrule
  \end{tabular}
  \vspace{-0.5em}
\end{table}

\partitle{API Cost}
We additionally report the API cost of collecting the 2,000 pairwise judgments with different proprietary LLM backbones. The reported cost covers the complete TSQAgent pipeline, including the Perceiver, Inspector, and Adjudicator, together with external Python tool calls and ReAct-style reasoning loops. As summarized in Table~\ref{tab:api_cost}, the annotation process consumes approximately 5.8--6.4 million input tokens and 0.26--0.35 million output tokens, with the resulting API cost depending on the selected LLM backbone.

\begin{table}[!htbp]
  \vspace{-0.5em}
  \centering
  \caption{Token consumption and API cost for collecting 2,000 pairwise quality judgments using different proprietary LLM backbones.}
  \label{tab:api_cost}
  \small
  \setlength{\tabcolsep}{8pt}
  \begin{tabular}{l|ccc}
    \toprule
    \textbf{LLM} & \textbf{Input Tokens} & \textbf{Output Tokens} & \textbf{API Cost} \\
    \midrule
    GPT-4o mini & 6.05M & 0.26M & \$1.08 \\
    Gemini 2.5 Flash & 5.82M & 0.28M & \$2.55 \\
    Claude Haiku 4.5 & 6.44M & 0.35M & \$7.50 \\
  \bottomrule
  \end{tabular}
  \vspace{-0.5em}
\end{table}

\section{Conclusion}
\label{conclu}

In this paper, we investigate the capability of LLMs for time series quality assessment by constructing a dedicated benchmark TSQBench and identifying two key limitations: unreliable quality dimension identification and weak quantitative comparison ability. To address these limitations, we propose TSQAgent, an agentic reasoning framework that devises reasoning-oriented dimension selection with tool-augmented quantitative analysis for reliable TS quality assessment. Experiments on TSQBench and eleven real-world datasets demonstrate that TSQAgent consistently improves TS quality reasoning and enables more effective quality-aware data selection.
Future work will address limitations related to generalization beyond synthetic quality degradations, predefined quality dimensions, agentic inference efficiency, and scalability to broader application settings, as detailed in Appendix~\ref{app:limitations}.

\clearpage

\begin{ack}
We would like to thank anonymous reviewers and area chairs for their constructive feedback and efforts in improving our paper. This work was supported by Tencent Rhino-Bird Research Fund (JR2026TEG604). This work was supported by the Open Research Fund of State Key Laboratory of Internet Architecture. We also gratefully acknowledge the computational resources and support provided by the National Supercomputer Center in Guangzhou.
\end{ack}

{
\small
\bibliography{conference}
\bibliographystyle{conference}
}


\newpage
\appendix

\section*{Appendix}
This appendix provides additional implementation details, experimental results, and analyses of TSQAgent. Section~\ref{rw} reviews related work on time series data quality assessment and agent-based methods for time series analysis. Section~\ref{app:benchmark} describes the construction and evaluation protocol of TSQBench. Section~\ref{app:grpo} presents the GRPO-based training details of the Perceiver, while Section~\ref{app:tools} documents the analytical tools used by the Inspector. Section~\ref{app:rating_model} describes the rating model architecture, training strategies, and supplementary evaluations. Section~\ref{appen:expd} provides experimental settings, additional results, and analyses. Section~\ref{app:case_study} presents qualitative case studies, followed by discussions of limitations in Section~\ref{app:limitations} and broader social impacts in Section~\ref{app:broader_impacts}. Finally, our code is available at \url{https://github.com/clsr1008/TSQAgent}.

\section{Related Works}
\label{rw}

\subsection{Time Series Data Quality Assessment}
Time series data quality assessment has been widely studied in the literature. Early approaches mainly rely on statistical feature analysis and heuristic rules to detect low-quality samples, such as threshold-based anomaly detection~\citep{clark2018adaptive}, distributional checks~\citep{tanaka2017time}, and manually designed quality indicators~\citep{wang2019time}. While these methods are simple and interpretable, they often fail to capture complex temporal dependencies and structural distortions in real-world time series.
Another line of research adapts data attribution techniques to time series for quality estimation, and can be broadly categorized into Influence Function-based~\citep{hampel1974influence} and Shapley value-based~\citep{shapley1953value} approaches. For influence-based methods, TimeInf~\citep{zhang2025timeinf} proposes a time-aware influence function to capture temporal dependencies, while ChInf~\citep{wang2024channel} extends this idea to multivariate settings via channel-level gradient approximations.
LTSV~\citep{wu2026lightweight} further leverages in-context finetuning to approximate influence functions for efficient data valuation on time series foundation models, incorporating temporal block aggregation to preserve temporal dependencies.
For Shapley-based methods, WindowSHAP~\citep{nayebi2023windowshap} introduces window-based decomposition to enable tractable attribution over long sequences. Additionally, ShapTST~\citep{cheng2025unifying} integrates Shapley computation into Transformer training to jointly support forecasting and explanation. Despite their theoretical soundness, these methods are computationally expensive, as the former requires intensive Hessian and gradient computations, while the latter suffers from exponential combinatorial complexity.
Recent studies have also investigated data-centric optimization for time series foundation models (TSFMs), including corpus balancing, distribution alignment, and data scaling~\citep{li2026quals,gong2026bridging,yao2025towards}. These efforts highlight the importance of training data characteristics for time series modeling, complementing sample-level quality assessment and selection.
More recently, motivated by the strong reasoning capabilities of large language models (LLMs), several works begin to explore LLM-based time series quality assessment~\citep{lan2025axis,cai2024timeseriesexam}. Among them, TSRating~\citep{wu2026rating} is a representative approach that defines multiple human-designed quality dimensions (e.g., trend, frequency, amplitude, and pattern consistency) and formulates pairwise comparison tasks to train a learned quality scoring model via Bradley--Terry optimization. This line of work demonstrates the potential of LLMs for semantic-level understanding of time series quality, but still heavily depends on manually specified dimensions and text-based reasoning, limiting its ability to perform fine-grained and quantitative quality assessment.

\subsection{Agent-based Methods for Time Series Analysis}
Recent advances have explored agentic frameworks for time series analysis, aiming to decompose complex workflows into modular reasoning and decision-making processes~\citep{cheng2026position}. A line of work focuses on automating end-to-end forecasting pipelines via multi-agent collaboration~\citep{shen2026metacaster}. For example, TimeSeriesScientist~\citep{zhao2025timeseriesscientist} introduces a four-agent system that covers data curation, model planning, forecasting, and reporting, while TimeCopilot~\citep{garza2025timecopilot} integrates multiple time series foundation models (TSFMs) with LLMs to automate feature analysis, model selection, and prediction. Similarly, AD-AGENT~\citep{yang2025ad} extends this paradigm to anomaly detection by coordinating agents for data preparation, model selection, and code generation.
Another line of research emphasizes reasoning-centric agents that iteratively refine predictions or analysis through structured reasoning and interaction. AlphaCast~\citep{zhang2025alphacast} formulates forecasting as an interactive co-reasoning process between human knowledge and LLMs, incorporating multi-source context and reflective optimization. TimeXL~\citep{jiang2025timexl} and TimeART~\citep{wu2026timeart} further adopt multi-agent or multi-stage reasoning loops to improve prediction accuracy and interpretability, while TS-Agent~\citep{liu2025ts} focuses on iterative evidence gathering and reasoning over raw time series signals.
In addition, several works explore tool-augmented and reinforcement learning-based agent designs for time series modeling. For instance, Cast-R1~\citep{tao2026cast} reformulates forecasting as a sequential decision-making process, where an agent interacts with external tools and is optimized via reinforcement learning to iteratively refine predictions.
Despite these advances, existing time series agents are primarily designed for forecasting, anomaly detection, or question answering, and focus on improving predictive performance or interpretability. In contrast, the problem of \emph{time series quality assessment}, particularly identifying decision-relevant quality dimensions and performing grounded, quantitative comparison, remains largely underexplored in agentic settings.

\section{TSQBench: Benchmark Construction and Evaluation Protocol}
\label{app:benchmark}

\subsection{Synthetic Time Series Template Generation}
\label{app:benchmark_template}

Each benchmark instance is constructed from a shared template time series, which serves as the common basis for pairwise time series. The template is designed to capture diverse yet realistic temporal structures, including trend, periodicity, and stochastic variations. Two correlated series $(\mathrm{series}_A, \mathrm{series}_B)$ are generated by duplicating the template, ensuring structural comparability and facilitating controlled dimension-specific degradations in subsequent steps.
Specifically, each template is constructed by composing three fundamental components: a trend term, a seasonal (periodic) term, and a stochastic noise term.

To enhance diversity, multiple composition strategies are adopted, including additive combinations, multiplicative interactions between trend and seasonality, and sequential compositions that concatenate multiple segments with distinct local patterns. The trend component covers a range of forms such as linear, piecewise, exponential, and logarithmic variations, while the seasonal component includes different periodic patterns (e.g., sinusoidal, square, and mixed harmonics). The noise component introduces low-level stochastic variations with different temporal structures (e.g., independent, autocorrelated, or time-varying).

All components are sampled from continuous parameter ranges and combined with randomized scaling and shifting, resulting in a broad distribution of temporal dynamics. Importantly, the baseline noise level is intentionally kept low, as dimension-specific degradations are introduced separately in later stages. This design ensures that the template captures clean underlying structures while allowing controlled manipulation of quality factors during benchmark construction.

\begin{table}[htbp!]
\centering
\caption{Quality dimension schema used in our work.}
\small
\setlength{\tabcolsep}{4pt}
\renewcommand{\arraystretch}{1.1}
\begin{tabular}{l p{10cm}}
\toprule
\textbf{Dimension} & \textbf{Description} \\
\midrule
\textbf{Missing Value} & Fraction of missing or unobserved values; higher ratios indicate lower quality. \\
\textbf{Noise Level} & Degree of noise or stochastic perturbations in the signal; higher levels indicate lower quality. \\
\textbf{Rare Pattern} & Anomalous observations caused by data defects (e.g., sensor glitches, corruption); more frequent or severe outliers indicate lower quality. \\
\textbf{Trend} & Clear and sustained directional movement over time; absence of stable direction (e.g., flat segments, erratic reversals) indicates low quality. \\
\textbf{Frequency} & Regular and stable oscillatory patterns over time; irregular or inconsistent cycles, such as varying periods or missing repetitions, indicate low quality. \\
\textbf{Amplitude} & Consistency and stable oscillation magnitude across cycles; varying or drifting swing strength over time indicates low quality. \\
\textbf{Pattern Consistency} & Global structural coherence over time; fragmented or unstable behavior, such as abrupt shifts or discontinuous transitions indicates low quality. \\
\bottomrule
\end{tabular}
\label{tab:quality_dimension}
\end{table}

\subsection{Active Quality Dimension Sampling}
\label{app:benchmark_dimension_sampling}

For each benchmark instance, a subset of dimensions defined in Table~\ref{tab:quality_dimension} is sampled to be active, where quality differences in the corresponding dimensions are introduced by degrading only one side (i.e., either $\mathrm{series}_A$ or $\mathrm{series}_B$).
We adopt a skewed distribution over the number of active dimensions in each benchmark instance to reflect real-world scenarios where quality differences are typically driven by a small number of dominant factors. The detailed distribution is shown in Table~\ref{tab:active_dim_dist}.

\begin{table}[h]
\centering
\caption{Distribution of active dimensions per benchmark instance.}
\small
\setlength{\tabcolsep}{6pt}
\renewcommand{\arraystretch}{1.1}
\begin{tabular}{c c l}
\toprule
\textbf{Number of Active Dimensions} & \textbf{Probability} & \textbf{Description} \\
\midrule
1   & 35\% & Most common: single-dimension comparison \\
2   & 35\% & Common: two-dimension comparison \\
3   & 20\% & Moderate frequency \\
4   & 8\%  & Less frequent \\
5   & 2\%  & Rare cases \\
\bottomrule
\end{tabular}
\label{tab:active_dim_dist}
\end{table}

\subsection{Dimension-specific Defect Injection}
\label{app:benchmark_defect_injection}

For each selected dimension, one of the two series is randomly assigned as the degraded side, while the other remains unchanged. We then introduce controlled, dimension-specific perturbations to the selected series to create asymmetric quality differences, while preserving the underlying shared temporal structure.
The type of degradation is determined by the semantic meaning of each quality dimension, and is sampled from a predefined set of transformation operators~\citep{xie2025chatts}. This design ensures that each dimension corresponds to a distinct and interpretable failure mode, enabling fine-grained benchmark evaluation.


For each quality dimension, we define a set of defect variants, each corresponding to a specific transformation operator (e.g., missingness injection, noise amplification, trend distortion). A variant is randomly selected during injection, and its parameters are sampled from predefined ranges. This design introduces both structural diversity and controlled controllability within each dimension.
Table~\ref{tab:defect_semantics} summarizes the implemented operators for all seven dimensions. The table reports the transformation family and the parameter regimes used by the current generator.

\begin{table}[!htbp]
\centering
\caption{Dimension-specific defect operators used in the synthetic benchmark.}
\small
\setlength{\tabcolsep}{4pt}
\renewcommand{\arraystretch}{1.18}
\begin{tabular}{>{\raggedright\arraybackslash}p{2.05cm}
                >{\raggedright\arraybackslash}p{3.25cm}
                >{\raggedright\arraybackslash}p{7.35cm}}
\toprule
\textbf{Dimension} & \textbf{Operators} & \textbf{Semantics and parameterization} \\
\midrule
\textbf{Missing Value}
& \texttt{random\_scatter}, \texttt{burst}, \texttt{periodic}
& Missing observations are inserted as isolated positions, contiguous dropout blocks, or approximately periodic gaps. The missing ratio is sampled from $[0.02,0.08]$ for scattered and periodic masks; burst masks use $1$--$3$ blocks with total burst ratio in $[0.02,0.05]$. This dimension is calibrated as heavy because the missing ratio is directly observable from the input statistics. \\
\textbf{Noise Level}
& \texttt{gaussian}, \texttt{heteroscedastic}, \texttt{impulsive}
& Global Gaussian noise increases the whole-series residual variance. Heteroscedastic noise affects one contiguous segment, with affected length ratio $[0.10,0.20]$ for light cases and $[0.30,0.70]$ for heavy cases. Impulsive noise corrupts short bursts. Noise multipliers are smaller for light cases ($1.2$--$2.0$ for Gaussian/heteroscedastic; $1.5$--$2.0$ for impulsive) and larger for heavy cases ($2.5$--$4.0$ or $3.0$--$5.0$). \\
\textbf{Rare Pattern}
& \texttt{point\_outlier}, \texttt{contextual}, \texttt{level\_shift}
& Point outliers add one to three isolated spikes with magnitude $2.5$--$3.5$ empirical standard deviations and are calibrated as heavy. Contextual anomalies create short V-shaped excursions of duration $2$--$4$ and magnitude $4.0$--$5.0$ standard deviations; these are treated as light because they are localized and shape-dependent. Temporary level shifts last $5$--$10$ steps with magnitude $1.5$--$2.5$ standard deviations and are calibrated as heavy. \\
\textbf{Trend}
& \texttt{flatten}, \texttt{drift}, \texttt{reversal}
& Trend defects modify a local structural component rather than the whole signal. Flattening replaces a segment of length ratio $[0.15,0.40]$ by its local mean plus amplified residual noise. Drift adds a gradual slope change across a long segment spanning from the early part of the series to the later part. Reversal reverses a segment of length ratio $[0.15,0.40]$ and level-aligns the boundary. These operators are calibrated as light because the degraded trend often preserves plausible local values. \\
\textbf{Frequency}
& \texttt{competing}, \texttt{jitter}, \texttt{period\_shift}
& Frequency defects operate on the oscillatory component using the known base period when available. Competing-frequency defects add one weak non-harmonic sinusoid in light cases, or two to three stronger components in heavy cases. Jitter applies smooth cumulative phase noise with coefficient of variation $[0.08,0.22]$. Period shifts modify the effective period in the last $20$--$50\%$ of the series by a factor in $[0.50,0.80]$. \\
\textbf{Amplitude}
& \texttt{random\_scale}, \texttt{decay}, \texttt{clip}
& Amplitude defects separate the baseline from the oscillatory component before perturbation. Random scaling samples per-cycle log-normal scale factors with coefficient of variation $[0.20,0.60]$ for light cases and $[0.80,1.10]$ for heavy cases. Decay applies a monotone envelope that either grows or shrinks the oscillatory magnitude toward a factor in $[0.20,0.70]$. Clipping caps oscillatory peaks and troughs at $40$--$80\%$ of their original maximum magnitude. \\
\textbf{Pattern Consistency}
& \texttt{variance\_switching}, \texttt{structural\_break}, \texttt{flat\_spots}, \texttt{mean\_drift}
& Consistency defects disturb global structural coherence. Variance switching alternates calm and volatile windows and is calibrated as heavy. Structural breaks introduce one or more persistent level shifts in the middle portion of the series. Flat spots replace random spans, typically $5$--$15\%$ of the sequence length, with a constant value. Mean drift adds a smoothed random-walk displacement to the signal level, with step scale proportional to the series standard deviation. \\
\bottomrule
\end{tabular}
\label{tab:defect_semantics}
\end{table}

\subsection{Evaluation Tasks and Metrics}

Building on the above design, which explicitly decomposes time series quality understanding into dimension identification and dimension-specific quality comparison reasoning, we evaluate models along two corresponding capabilities on a set of 1,000 synthetic samples. 
For dimension identification, given a pair $(\mathrm{series}_A, \mathrm{series}_B)$, the model is required to select a subset of relevant dimensions $\mathcal{D}_{pred}$ from the full dimension set $\mathcal{D}$, where the ground-truth set $\mathcal{D}_{gt}$ denotes the sampled active dimensions exhibiting quality differences between the two series.
For quality comparison, the model is provided with the ground-truth active dimension and asked to determine which series exhibits higher quality under the specified dimension, where the ground-truth preference corresponds to the unperturbed series during defect injection. 
Performance is measured via Precision, Recall for dimension identification, and comparison accuracy for the quality comparison.

\section{Implementation Details of GRPO-based Perceiver Training}
\label{app:grpo}

\textbf{Training Target and Policy Parameterization.} We fine-tune \texttt{Qwen3-4B} as the Perceiver policy using the TRL implementation of GRPO. The trained policy is parameterized with LoRA rather than full-model updates. In the default configuration, LoRA adapters are inserted into the attention projection layers \texttt{q\_proj}, \texttt{k\_proj}, and \texttt{v\_proj}, with rank $r=16$, scaling factor $\alpha=32$, dropout $0.05$, no bias terms, and causal-language-model task type. This keeps the trainable parameter budget small while adapting the attention subspace most directly involved in Perceiver planning.

The Perceiver is trained to output a JSON object containing a short perception summary and a list-valued \texttt{planned\_dimensions} field. Only the \texttt{planned\_dimensions} field is used by the GRPO reward. Thus, the implemented RL objective optimizes dimension selection and output validity, not the Inspector's later tool-use decision. This distinction is important because the final TSQAgent pipeline still delegates detailed measurement and tool invocation to the Inspector, while the tailored GRPO is used to improve the Perceiver's planning precision.

\textbf{Training Data.} We use the synthetic time series quality benchmark as the supervision source for GRPO. Each training instance provides a pair of time series segments and a ground-truth target dimension set $D_{\mathrm{gt}}$, which indicates the quality dimensions that should be selected by the Perceiver. The data construction process, defect taxonomy, label semantics, and split details are described in Appendix~\ref{app:benchmark}.

\textbf{Reward Function.} The proposed TS quality-tailored GRPO reward combines a small format-compliance term with a dominant dimension-selection term. The format term encourages the model to return parseable JSON following the Perceiver schema, while the dimension term measures the precision of the predicted \texttt{planned\_dimensions} against the target dimension set $D_{\mathrm{gt}}$. The effective reward used by the trainer is
\begin{equation}
R = 0.10\,R_{\mathrm{format}} + 0.90\,R_{\mathrm{dim}},
\end{equation}
where $R_{\text{format}}$ ensures valid structured output, and $R_{\text{dim}}$ measures the alignment between predicted and ground-truth dimensions:
\begin{equation}
R_{\text{dim}} = \frac{|\mathcal{D}_{pred} \cap \mathcal{D}_{gt}|}{|\mathcal{D}_{pred}|}.
\end{equation}
The 0.1/0.9 weighting reflects the role of GRPO in this project: valid structured output is necessary, but the main objective is to make the Perceiver select relevant dimensions without over-selecting. This precision-oriented design discourages forwarding redundant dimensions to the Inspector, thereby reducing downstream tool calls and irrelevant evidence. 

\textbf{Tailored GRPO Optimization Configuration.} For each prompt, GRPO samples $K=8$ completions from the current policy using temperature $1.0$. These completions form the comparison group for relative policy optimization. With the default gradient accumulation of four steps, each optimizer update aggregates four unique prompts and 32 sampled completions per GPU. Under DDP with $N$ GPUs, the effective update aggregates $4N$ prompts and $32N$ completions.

The training run uses one epoch, learning rate $10^{-5}$, cosine learning-rate decay, and warmup ratio $0.10$. The KL regularization coefficient is $\beta=0.04$, which constrains the fine-tuned policy against the reference behavior used internally by TRL's GRPO trainer. This is important because the reward is sparse and format-sensitive: without KL pressure, the policy could overfit to short, reward-hacking completions or collapse into degenerate JSON templates. The maximum generated completion length is 300 tokens, which is sufficient for the required Perceiver JSON while limiting long reasoning traces.

\textbf{Runtime and Checkpointing.} The model is loaded in bf16 by default, uses the PyTorch SDPA attention implementation, and enables gradient checkpointing to reduce memory usage during GRPO sampling. The default full-training uses one prompt per GPU, eight generations per prompt, and four gradient-accumulation steps. The trainer saves LoRA checkpoints at the end of each epoch and limits the number of saved checkpoints to the number of epochs. In the full TSQAgent pipeline, the LoRA-served model can be assigned only to the Perceiver, while the Inspector and Adjudicator continue to use the base model endpoint; this isolates the effect of GRPO to the planning stage. Our training is conducted on a single NVIDIA A100 GPU.

\section{Implementation Details of Time Series Analytical Tools}
\label{app:tools}
This section describes the implementation of the 16 analytical tools used by the Inspector in TSQAgent. The tools are organized into three categories: data completeness and noise analysis, outlier and rare pattern detection, and temporal structure and pattern analysis. Each tool is accessed through a unified registry, operates on a specified time series (\texttt{series\_name} $\in \{\mathrm{A}, \mathrm{B}\}$), and returns structured quantitative evidence for the Inspector's reasoning process.

All tools operate on a univariate time series $x = \{x_t\}_{t=1}^T$. Tools based on summary statistics ignore missing entries when computing valid statistics, while tools requiring continuous temporal structure, such as autocorrelation and trend analysis, fill missing values via linear interpolation before computation.

\subsection{Data Completeness and Noise Analysis Tools}

\textbf{(1) Missing Ratio.}
The \texttt{missing\_ratio} tool quantifies data completeness by counting the fraction of missing entries:
\begin{equation}
\texttt{missing\_ratio}(x) = \frac{|\{x_t \mid x_t = \text{NaN}\}|}{T}.
\end{equation}
It returns \texttt{missing\_ratio}, where larger values indicate more severe data loss. This tool is the primary evidence source for the \textit{Missing Value} dimension.

\textbf{(2) Noise Profile.}
The \texttt{noise\_profile} tool estimates the high-frequency component of a series by subtracting a rolling mean from the valid observations:
\begin{equation}
\texttt{noise\_ratio}(x) = 
\frac{\sigma(x - \hat{x}_{\mathrm{roll}}^{(w)})}{\sigma(x)},
\end{equation}
where $\hat{x}_{\mathrm{roll}}^{(w)}$ is a moving average with window size $w$. If $w$ is not specified, the implementation adapts it to the sequence length. The tool returns \texttt{noise\_std}, \texttt{signal\_std}, \texttt{noise\_ratio}, \texttt{noise\_type}, and \texttt{window\_used}. The noise type is determined by the lag-1 autocorrelation of the valid series, where strongly autocorrelated residual behavior is labeled as red noise and weakly autocorrelated behavior as white noise.

\textbf{(3) Volatility.}
The \texttt{volatility} tool measures local instability using the rolling standard deviation of first-order differences:
\begin{equation}
\texttt{volatility}(x) =
\frac{1}{N_w}\sum_i \mathrm{Std}\left(\Delta x_{i:i+w}\right),
\quad \Delta x_t = x_t - x_{t-1}.
\end{equation}
It returns \texttt{mean\_volatility}, \texttt{max\_volatility}, and \texttt{window\_used}. While \texttt{noise\_profile} measures residual variation after smoothing, \texttt{volatility} captures abrupt step-to-step changes, making it a complementary tool for the \textit{Noise Level} dimension.

\textbf{(4) Range Statistics.}
The \texttt{range\_stats} tool computes local statistics over an interval $[i,j)$:
\begin{equation}
\texttt{range\_stats}(x, i, j, s) = s(x_{i:j}),
\quad s \in \{\mathrm{mean}, \mathrm{std}, \mathrm{max}, \mathrm{min}, \mathrm{sum}\}.
\end{equation}
It returns the selected statistic value together with the clipped interval and segment length. This tool is used as local supporting evidence, for example after a change point has been identified or when the Inspector needs to compare a specific region rather than the entire series.

\subsection{Outlier and Rare Pattern Detection Tools}

\textbf{(5) Z-score Outliers.}
The \texttt{zscore\_outlier} tool detects global point anomalies using standard deviation thresholding:
\begin{equation}
z_t = \frac{|x_t - \mu|}{\sigma}, \quad
\mathbb{I}_{\mathrm{outlier}}(t)=\mathbb{I}[z_t > \lambda].
\end{equation}
It returns \texttt{anomaly\_count}, \texttt{anomaly\_ratio}, \texttt{anomaly\_indices}, \texttt{anomaly\_values}, and \texttt{threshold\_used}. This tool provides a simple first-pass detector for isolated spikes, especially when the series is approximately stationary.

\textbf{(6) Outlier Density.}
The \texttt{outlier\_density} tool estimates anomaly density using interquartile-range fences:
\begin{equation}
\mathrm{IQR} = Q_3 - Q_1,\quad
\ell = Q_1 - 1.5\,\mathrm{IQR},\quad
u = Q_3 + 1.5\,\mathrm{IQR}.
\end{equation}
Points outside $[\ell,u]$ are counted as outliers. The tool returns \texttt{outlier\_count}, \texttt{outlier\_ratio}, \texttt{iqr}, \texttt{lower\_fence}, and \texttt{upper\_fence}. Compared with Z-score thresholding, this detector is more robust to heavy-tailed distributions because it relies on quantiles rather than mean and variance.

\textbf{(7) MAD-based Residual Outliers.}
The \texttt{mad\_residual\_outlier} tool first removes local trend using a rolling mean and then applies median absolute deviation (MAD) scoring to the residuals:
\begin{equation}
r_t = x_t - \hat{x}_{\mathrm{roll},t}^{(w)}, \quad
m_t = \frac{0.6745\,|r_t - \mathrm{median}(r)|}{\mathrm{MAD}(r)}.
\end{equation}
Points with $m_t > \lambda$ are flagged as anomalies. The tool returns \texttt{anomaly\_count}, \texttt{anomaly\_ratio}, \texttt{anomaly\_indices}, \texttt{anomaly\_values}, \texttt{mad}, \texttt{threshold\_used}, and \texttt{window\_used}. This robust detector is preferred when a series has trend or seasonal drift, where global Z-scores may produce false positives.

\textbf{(8) Contextual Rare Pattern.}
The \texttt{contextual\_rare\_pattern} tool identifies points that deviate from their local context. For each time index, it fits a linear model over the preceding context window and scores the prediction error:
\begin{equation}
\hat{x}_t = a_t t + b_t,\quad
e_t = |x_t - \hat{x}_t|.
\end{equation}
The errors are normalized using MAD, and points with unusually large local prediction errors are flagged. The tool returns \texttt{anomaly\_count}, \texttt{anomaly\_ratio}, \texttt{anomaly\_indices}, \texttt{anomaly\_values}, \texttt{threshold\_used}, and \texttt{context\_window\_used}. This tool is useful for contextual events such as sudden drops, V-shaped deviations, or local structural breaks that may not be extreme under global statistics.

\subsection{Temporal Structure and Pattern Analysis Tools}

This category corresponds to tools for assessing structural richness and temporal organization, covering the \textit{Trend}, \textit{Frequency}, \textit{Amplitude}, and \textit{Pattern Consistency} dimensions.

\textbf{(9) Trend Classifier.}
The \texttt{trend\_classifier} tool estimates global trend direction and strength via linear regression:
\begin{equation}
x_t = \beta_0 + \beta_1 t + \epsilon_t,\quad
\texttt{trend\_strength}=R^2.
\end{equation}
It returns \texttt{direction}, \texttt{slope}, \texttt{trend\_strength}, \texttt{segment\_clarity}, and \texttt{segment\_count}. In addition to the global fit, the implementation uses change points to segment the series and computes per-segment trend clarity, which captures whether each natural segment exhibits a coherent direction even when the overall series is not globally linear.

\textbf{(10) Change Point Detection.}
The \texttt{change\_point\_detector} tool detects structural breakpoints where the mean level or temporal behavior changes:
\begin{equation}
\texttt{change\_point\_detector}(x)=\{\tau_1,\tau_2,\ldots,\tau_K\}.
\end{equation}
The implementation uses the PELT algorithm with an $\ell_2$ cost when the \texttt{ruptures} package is available, and falls back to a CUSUM-style detector otherwise. It returns \texttt{change\_point\_count}, \texttt{change\_point\_indices}, and \texttt{method}. This tool supports trend, amplitude, and pattern-consistency analysis by localizing regions where the series changes character.

\textbf{(11) Seasonality Detection.}
The \texttt{seasonality\_detector} tool searches for dominant periodicity using the autocorrelation function:
\begin{equation}
\rho(\tau)=
\frac{\mathbb{E}[(x_t-\bar{x})(x_{t+\tau}-\bar{x})]}{\mathrm{Var}(x)},\quad
\tau^\star = \arg\max_{\tau} \rho(\tau).
\end{equation}
The implementation detects significant ACF peaks with prominence filtering and suppresses harmonics so that near-integer multiples of a stronger base period are not double-counted. It returns \texttt{dominant\_period}, \texttt{seasonal\_strength}, \texttt{top\_periods}, \texttt{dominance\_ratio}, and \texttt{peak\_count}. This is the primary tool for the \textit{Frequency} dimension.

\textbf{(12) Autocorrelation.}
The \texttt{autocorr} tool computes dependency at a specified lag:
\begin{equation}
\texttt{autocorr}(x,\tau)=\mathrm{corr}(x_{1:T-\tau},x_{1+\tau:T}).
\end{equation}
It returns \texttt{lag} and \texttt{autocorrelation}. Unlike \texttt{seasonality\_detector}, which searches across many lags, this tool verifies a specific lag hypothesis and is therefore used as supporting evidence for frequency and smoothness-related judgments.

\textbf{(13) Pattern Consistency Indicators.}
The \texttt{pattern\_consistency\_indicators} tool computes a collection of structural coherence measures:
\begin{align}
\texttt{lumpiness} &= \mathrm{Var}\left(\{\mathrm{Var}(x_{W_k})\}_k\right),\\
\texttt{roughness} &= \frac{1}{T-1}\sum_{t=1}^{T-1}|x_{t+1}-x_t|.
\end{align}
It also returns \texttt{flat\_ratio}, \texttt{longest\_flat\_ratio}, \texttt{crossing\_points}, and \texttt{crossing\_rate}. These indicators capture bursty variance, stagnant intervals, frequent mean crossings, and jaggedness. They are the primary evidence source for the \textit{Pattern Consistency} dimension.

\textbf{(14) Stationarity Test.}
The \texttt{stationarity\_test} tool provides statistical tests for stationarity. It supports the Augmented Dickey--Fuller (ADF) test and the KPSS test:
\begin{equation}
\texttt{stationarity\_test}(x)=(\mathrm{statistic}, p, \mathrm{is\_stationary}).
\end{equation}
For ADF, the null hypothesis is that the series has a unit root, so $p<0.05$ indicates stationarity. For KPSS, the null hypothesis is stationarity, so $p>0.05$ indicates stationarity. The tool returns \texttt{test}, \texttt{statistic}, \texttt{p\_value}, and \texttt{is\_stationary}. It is used as supporting evidence for trend and pattern-consistency analysis.

\textbf{(15) Rolling Amplitude.}
The \texttt{rolling\_amplitude} tool computes local amplitude as the range within a sliding window:
\begin{equation}
\texttt{local\_range}_i = \max(x_{i:i+w}) - \min(x_{i:i+w}).
\end{equation}
It returns \texttt{mean\_local\_range}, \texttt{cv\_local\_range}, \texttt{max\_local\_range}, \texttt{min\_local\_range}, and \texttt{window}. This tool is a general-purpose amplitude measure and is especially useful when a series is not clearly oscillatory.

\textbf{(16) Cycle Amplitude.}
The \texttt{cycle\_amplitude} tool measures cycle-level amplitude consistency for oscillatory series. It first performs a gate check requiring at least two significant peaks and two significant troughs, where significance is determined by peak prominence relative to the series standard deviation. It then pairs adjacent extrema and computes peak-to-trough magnitudes:
\begin{equation}
a_k = |x_{p_k} - x_{q_k}|,\quad
\texttt{amplitude\_cv}=\frac{\sigma(\{a_k\})}{\mu(\{a_k\})}.
\end{equation}
The tool returns \texttt{oscillatory}, \texttt{cycle\_count}, \texttt{mean\_amplitude}, \texttt{amplitude\_cv}, \texttt{amplitude\_trend}, \texttt{peak\_count}, and \texttt{trough\_count}. It is the primary amplitude tool when the series has clear oscillations; otherwise, the Inspector falls back to \texttt{rolling\_amplitude}.

\section{Rating Model Design and Training Strategies}
\label{app:rating_model}


\subsection{Rating Model Architecture and Training Objective}
\label{app:model_arch}
The rating model is designed to assign a scalar quality score to each time series segment. Given an input time series $\mathbf{x}$, the model outputs a scalar score reflecting its quality under predefined criteria. The architecture follows a two-stage design: representation extraction and quality mapping.

\textbf{Representation Learning.}
We adopt a pretrained time series foundation model, MOMENT~\citep{goswami2024moment}, to encode input series into latent representations. MOMENT contains approximately 109M parameters and is pretrained to capture diverse temporal patterns, providing a strong feature extractor. To ensure stable and efficient training, the encoder is kept frozen during optimization.

\textbf{Quality Mapping.}
The extracted representation is fed into a lightweight multi-layer perceptron (MLP) to produce a scalar quality score. The MLP consists of three fully connected layers with hidden dimension 256. Each layer is followed by LayerNorm and ReLU activation, and residual connections are incorporated to improve optimization stability.

The model is trained using pairwise preference data derived from agent judgments. Specifically, given a pair of time series $(\mathbf{x}_i, \mathbf{x}_j)$ with preference label $p_{i \succ j}$ indicating whether $\mathbf{x}_i$ is preferred over $\mathbf{x}_j$, the loss is defined as:
\begin{equation}
\label{eq:rating_loss}
\mathcal{L}_\theta = 
\mathbb{E}_{(\mathbf{x}_i, \mathbf{x}_j, p_{i \succ j}) \in \mathcal{J}}
\Big[
- p_{i \succ j} \log \sigma\big(s_\theta(\mathbf{x}_i) - s_\theta(\mathbf{x}_j)\big)
- (1 - p_{i \succ j}) \log \sigma\big(s_\theta(\mathbf{x}_j) - s_\theta(\mathbf{x}_i)\big)
\Big],
\end{equation}
where $\mathcal{J}$ denotes the set of pairwise comparisons produced by the agent, and $s_\theta(\mathbf{x})$ is the predicted quality score parameterized by $\theta$. This formulation corresponds to a standard pairwise ranking objective and is consistent with the maximum likelihood estimation.

\subsection{Pairwise Comparison Sampling Details}
\label{app:pair_sampling}

For the main experiments in Section~\ref{subsec:select}, we train a separate rating model for each dataset. Given a training pool containing $N$ samples, we uniformly sample $N/2$ pairs at random without replacement for pairwise quality assessment, balancing training-set coverage and annotation efficiency. The resulting preference judgments are used as supervision for training the corresponding rating model. To reduce noisy supervision, we discard pairwise comparisons whose confidence scores estimated by the Adjudicator are below 0.5. Across all datasets, fewer than 10\% of the sampled comparisons are removed by this filtering procedure. Table~\ref{tab:pairwise_sampling} reports the exact number of sampled pairwise comparisons for each dataset before confidence filtering.

\begin{table}[!htbp]
    \centering
    \caption{Number of pairwise comparisons sampled for dataset-specific rating model training before confidence filtering.}
    \label{tab:pairwise_sampling}
    \small
    \setlength{\tabcolsep}{4pt}
    \begin{tabular}{l | c c c c c c c c c c c}
        \toprule
        \textbf{Dataset} 
        & Elec. & ExRate & Traffic & Weather 
        & M4Y & M4M & M4D 
        & MImg & CBF & BME & HW \\
        \midrule
        \textbf{Pairs}
        & 1342 & 1340 & 1343 & 1338 
        & 1391 & 1381 & 1379 
        & 503 & 435 & 435 & 500 \\
        \bottomrule
    \end{tabular}
\end{table}

For the cross-dataset TSFM fine-tuning experiments in Section~\ref{subsec:tsfm}, we instead train a shared rating model using the meta-learning strategy described in Appendix~\ref{app:meta_learning}. In this setting, we sample 500 pairwise comparisons from each dataset to construct the multi-domain training pool, allowing the rating model to leverage quality supervision across datasets and learn transferable quality representations.

\subsection{Effectiveness of Pairwise Quality Assessment}
\label{app:pairwise_effectiveness}

We adopt pairwise comparison to elicit more reliable quality assessments from LLMs. Directly assigning numerical quality scores to individual time series requires calibrating an absolute quality scale, whereas pairwise comparison only requires relative judgments between two samples, which are generally easier and more stable to obtain. Similar pairwise preference-based paradigms have also been adopted in recent quality evaluation studies, such as QuRating~\citep{wettig2024qurating}.

To further validate this design, we compare pairwise comparison with direct rating on the Weather dataset, where the two approaches differ in how quality judgments are elicited while following the same data selection pipeline with a 50\% selection budget. We evaluate their effectiveness using three downstream forecasting models (Linear, CNN, and PatchTST) in terms of forecasting RMSE. As shown in Table~\ref{tab:pairwise_vs_direct}, pairwise comparison consistently achieves lower forecasting RMSE across all three models, supporting the effectiveness of pairwise preference modeling for time series quality assessment.

\begin{table}[!htbp]
\centering
\caption{Comparison of pairwise comparison and direct rating on the Weather dataset. Results report forecasting RMSE ($\downarrow$) as mean $\pm$ standard deviation over 10 random seeds. Best mean results are \textbf{bolded}.}
\label{tab:pairwise_vs_direct}
\small
\setlength{\tabcolsep}{6pt}
\begin{tabular}{lccc}
\toprule
\textbf{Method} & \textbf{Linear} & \textbf{CNN} & \textbf{PatchTST} \\
\midrule
Pairwise Comparison
& $\mathbf{0.593} \pm 0.016$
& $\mathbf{0.723} \pm 0.015$
& $\mathbf{0.450} \pm 0.012$ \\
Direct Rating
& $0.607 \pm 0.020$
& $0.738 \pm 0.012$
& $0.455 \pm 0.010$ \\
\bottomrule
\end{tabular}
\end{table}

\subsection{Cross-Dataset Quality Scoring via Meta-Learning}
\label{app:meta_learning}
To enable consistent quality scoring across heterogeneous datasets, we further train the rating model using a meta-learning strategy. Unlike single-dataset training, where scores are only comparable within each dataset, meta-learning allows the model to learn a shared scoring function that generalizes across different data distributions.

In our setting, each dataset is treated as a separate task. We consider four datasets used in the main experiments (Electricity, ExchangeRate, Traffic, and Weather), and construct meta-training episodes by sampling tasks from this set. For each task, the available pairwise preference data is split into a support set and a query set. The model is first adapted to the support set with a small number of gradient updates, and then evaluated on the query set.

The meta-objective optimizes the model parameters such that the adapted model performs well on the query set across all tasks:
\begin{equation}
\small
\min_{\theta} \sum_{\mathcal{T}_i}
\mathcal{L}_{\mathcal{T}_i}^{\text{query}}
\Big(
\theta - \alpha \nabla_{\theta} \mathcal{L}_{\mathcal{T}_i}^{\text{support}}(\theta)
\Big),
\end{equation}
where $\theta$ denotes the model parameters and $\alpha$ is the inner-loop learning rate. This procedure encourages the model to capture dataset-invariant notions of data quality, making the resulting scores comparable across datasets.

In practice, we adopt a first-order approximation for efficiency and perform a small number of inner-loop updates. The resulting meta-trained rating model is then used to assign quality scores over the merged data pool for cross-dataset selection.

\section{Additional Experimental Details, Results, and Analyses}
\label{appen:expd}

\subsection{Benchmark Datasets and Baselines}
\label{app:configs}

\partitle{Experimental Setting and Comparability}
TSQAgent and TSRating~\citep{wu2026rating} both investigate LLM-based quality assessment for time series data. To facilitate direct and controlled comparisons, we adopt part of the evaluation protocol used in TSRating, particularly its benchmark datasets and data valuation baselines. Maintaining a consistent experimental setting helps isolate the contributions of the proposed agentic framework from differences in dataset selection or baseline configurations. Beyond these shared settings, TSQAgent extends the evaluation scope to examine additional capabilities, including cross-dataset data selection for time series foundation model fine-tuning (Section~\ref{subsec:tsfm}), tool-augmented reasoning (Section~\ref{subsec:ablation}), and interpretability through intermediate dimension-level evidence (Appendix~\ref{app:dimension_interpretability}).

\partitle{Benchmark Datasets}
Our experiments involve eleven benchmark datasets spanning different temporal characteristics and application domains, covering both forecasting and classification tasks. The forecasting benchmarks include four widely used datasets and three subsets of the M4 competition, while the classification benchmarks comprise four datasets with diverse temporal patterns.

\begin{itemize}

\item \textit{Electricity}~\citep{trindade2015electricityloaddiagrams20112014}:
Hourly electricity usage measurements, expressed in kWh, are available for 321 customers over the period from 2012 to 2014. For univariate forecasting, we extract the consumption history associated with \texttt{MT\_320} as the target series.

\item \textit{ExchangeRate}~\citep{lai2018modeling}:
This benchmark tracks the daily exchange rates of eight currencies from 1990 through 2016. Following the established univariate forecasting setting, we use the Singapore dollar exchange rate as the prediction target.

\item \textit{Traffic}~\citep{lai2018modeling}:
The traffic measurements consist of hourly freeway occupancy observations recorded by sensors in the San Francisco Bay Area during 2015--2016. Our experiments focus on the observations from sensor \texttt{861}.

\item \textit{Weather}~\citep{wu2021autoformer}:
Meteorological observations from approximately 1,600 locations across the United States form this benchmark, covering variables such as temperature, humidity, and wind speed. We conduct the univariate experiments using \texttt{wet\_bulb} as the target variable.

\item \textit{M4}~\citep{makridakis2018m4}:
The M4 forecasting competition provides a collection of 100,000 time series sampled at different temporal frequencies, including yearly, monthly, daily, weekly, and hourly intervals. We select the \textit{Yearly}, \textit{Monthly}, and \textit{Daily} subsets to evaluate data quality assessment across different forecasting frequencies.

\item \textit{MedicalImages}~\citep{dau2019ucr}:
This classification benchmark comprises more than 1,000 univariate sequences, each containing 99 observations derived from pixel intensity variations. The sequences are assigned to 10 categories representing different medical image classes.

\item \textit{CBF}~\citep{dau2019ucr}:
CBF consists of 930 synthetically generated univariate sequences, each with 128 time steps. Its three classes are characterized by distinct piecewise geometric patterns, providing a controlled setting for time series classification.

\item \textit{BME}~\citep{dau2019ucr}:
The BME benchmark contains 180 synthetic univariate sequences of length 128, organized into three classes distinguished by controlled variations in their temporal patterns.

\item \textit{Handwriting}~\citep{bagnall2018uea}:
This multivariate classification benchmark consists of 1,000 handwriting trajectories. Each trajectory is represented by three channels over 152 time steps and belongs to one of 26 character classes.

\end{itemize}

\partitle{Baselines}
We evaluate TSQAgent against several established approaches to data valuation and subset selection, encompassing Shapley-based valuation, out-of-bag estimation, influence-based attribution, and random sampling. These methods provide sample-level utility estimates or selection criteria that can be used to construct training subsets.

\begin{itemize}

\item \textbf{DataShapley}~\citep{ghorbani2019data}:
DataShapley applies the Shapley value principle to training data valuation. It estimates the importance of an individual sample by considering its marginal contribution to predictive performance across different training subsets.

\item \textbf{DataOob}~\citep{kwon2023data}:
DataOob evaluates training sample utility through out-of-bag predictions. By examining model behavior on samples excluded from particular training subsets, it estimates how individual observations contribute to predictive generalization.

\item \textbf{TimeInf}~\citep{zhang2025timeinf}:
TimeInf extends influence-based data attribution to temporal data. It estimates the effects of individual training samples on model predictions while accounting for dependencies within time series.

\item \textbf{TSRating}~\citep{wu2026rating}:
TSRating constructs a time series quality scoring function from pairwise preference supervision. It combines learned time series representations with a rating model to estimate scalar quality scores, allowing samples to be ranked according to their assessed quality.

\item \textbf{Random}:
Random selection serves as a reference strategy that constructs training subsets without considering sample quality. It provides a simple baseline for evaluating the benefits of quality-aware data selection.

\end{itemize}

\subsection{Variability of Experimental Results Across Random Seeds}
\label{app:std}

To assess the stability of the experimental results, we report the mean and standard deviation over 10 random seeds for the main data selection experiments in Table~\ref{tab:all_tasks} and the component ablation study in Section~\ref{subsec:ablation}. Overall, TSQAgent achieves competitive performance with relatively small variations across random seeds, further supporting the robustness of the reported improvements.

\partitle{Data Selection Performance}
We first report the detailed results of data selection methods with CNN as the downstream model under a 50\% high-quality data budget. Tables~\ref{tab:std_forecasting} and \ref{tab:std_classification} present the mean and standard deviation for forecasting and classification tasks, respectively.

\begin{table}[!htbp]
\centering
\caption{Mean $\pm$ standard deviation over 10 random seeds for the data selection experiments in Table~\ref{tab:all_tasks}, using CNN as the downstream model on forecasting datasets. Best mean results are \textbf{bolded}.}
\label{tab:std_forecasting}
\scriptsize
\setlength{\tabcolsep}{3pt}
\begin{tabular}{l|cccc|ccc}
\toprule
\textbf{Method} &
\multicolumn{4}{c|}{\textbf{Long-term (RMSE $\downarrow$)}} &
\multicolumn{3}{c}{\textbf{Short-term (MAPE $\downarrow$)}} \\
& Electricity & ExchangeRate & Traffic & Weather & M4 Yearly & M4 Monthly & M4 Daily \\
\midrule
Random
& $1.592 \pm 0.075$ & $1.474 \pm 0.073$ & $0.504 \pm 0.010$ & $0.769 \pm 0.033$
& $2.332 \pm 0.460$ & $1.124 \pm 0.042$ & $1.193 \pm 0.055$ \\
DataOob
& $1.609 \pm 0.049$ & $1.527 \pm 0.042$ & $0.552 \pm 0.011$ & $0.737 \pm 0.009$
& $2.957 \pm 0.552$ & $1.208 \pm 0.067$ & $1.346 \pm 0.085$ \\
DataShapley
& $1.529 \pm 0.060$ & $1.598 \pm 0.042$ & $0.475 \pm 0.012$ & $0.767 \pm 0.025$
& $2.553 \pm 0.548$ & $1.117 \pm 0.022$ & $1.159 \pm 0.036$ \\
TimeInf
& $1.530 \pm 0.066$ & $1.515 \pm 0.053$ & $0.505 \pm 0.007$ & $0.758 \pm 0.022$
& $2.289 \pm 0.334$ & $1.117 \pm 0.034$ & $1.103 \pm 0.044$ \\
TSRating
& $1.511 \pm 0.050$ & $1.429 \pm 0.036$ & $0.428 \pm 0.008$ & $0.734 \pm 0.025$
& $1.782 \pm 0.653$ & $\mathbf{1.075} \pm 0.056$ & $1.108 \pm 0.051$ \\
\rowcolor{cyan!10}
\textbf{TSQAgent}
& $\mathbf{1.419} \pm 0.054$ & $\mathbf{1.371} \pm 0.037$ & $\mathbf{0.412} \pm 0.009$ & $\mathbf{0.723} \pm 0.015$
& $\mathbf{1.744} \pm 0.354$ & $1.108 \pm 0.027$ & $\mathbf{1.096} \pm 0.031$ \\
\bottomrule
\end{tabular}
\end{table}

\begin{table}[!htbp]
\centering
\caption{Mean $\pm$ standard deviation over 10 random seeds for the data selection experiments in Table~\ref{tab:all_tasks}, using CNN as the downstream model on classification datasets. Best mean results are \textbf{bolded}.}
\label{tab:std_classification}
\footnotesize
\setlength{\tabcolsep}{5pt}
\begin{tabular}{l|cccc}
\toprule
\textbf{Method} &
\multicolumn{4}{c}{\textbf{Classification (Accuracy $\uparrow$)}} \\
& MedicalImages & CBF & BME & Handwriting \\
\midrule
Random
& $0.554 \pm 0.015$ & $0.595 \pm 0.061$ & $0.495 \pm 0.027$ & $0.151 \pm 0.009$ \\
DataOob
& $0.514 \pm 0.014$ & $0.663 \pm 0.016$ & $0.521 \pm 0.022$ & $0.155 \pm 0.006$ \\
DataShapley
& $0.550 \pm 0.011$ & $0.593 \pm 0.045$ & $0.564 \pm 0.028$ & $0.159 \pm 0.004$ \\
TimeInf
& $0.567 \pm 0.006$ & $0.563 \pm 0.035$ & $0.535 \pm 0.046$ & $0.155 \pm 0.007$ \\
TSRating
& $0.561 \pm 0.007$ & $\mathbf{0.679} \pm 0.018$ & $0.575 \pm 0.018$ & $0.159 \pm 0.006$ \\
\rowcolor{cyan!10}
\textbf{TSQAgent}
& $\mathbf{0.574} \pm 0.008$ & $0.668 \pm 0.019$ & $\mathbf{0.580} \pm 0.021$ & $\mathbf{0.162} \pm 0.005$ \\
\bottomrule
\end{tabular}
\end{table}

\partitle{Component Analysis}
We further report the mean and standard deviation over 10 random seeds for three ablation studies, including GRPO training, tool-augmented quantitative comparison, and quality dimension design. As shown in Tables~\ref{tab:std_grpo}, \ref{tab:std_tools}, and \ref{tab:std_dimensions}, the results generally support the effectiveness of GRPO training, tool-augmented quantitative comparison, and the proposed quality dimension design, with relatively small variations across random seeds.

\begin{table}[!htbp]
\centering
\caption{Mean $\pm$ standard deviation over 10 random seeds for the Perceiver ablation study in Figure~\ref{fig:grpo_ablation}(a). Results report forecasting RMSE ($\downarrow$).}
\label{tab:std_grpo}
\small
\setlength{\tabcolsep}{6pt}
\begin{tabular}{lcccc}
\toprule
\textbf{Method} & \textbf{Electricity} & \textbf{ExchangeRate} & \textbf{Traffic} & \textbf{Weather} \\
\midrule
w/o GRPO
& $0.449 \pm 0.018$ & $0.243 \pm 0.007$ & $0.360 \pm 0.005$ & $0.512 \pm 0.011$ \\
w/ GRPO
& $0.400 \pm 0.017$ & $0.218 \pm 0.007$ & $0.353 \pm 0.004$ & $0.496 \pm 0.010$ \\
\bottomrule
\end{tabular}
\end{table}

\begin{table}[!htbp]
\centering
\caption{Mean $\pm$ standard deviation over 10 random seeds for the tool-augmented quantitative comparison ablation study in Figure~\ref{fig:tool_ablation}, using Qwen3-4B. Results report forecasting RMSE ($\downarrow$).}
\label{tab:std_tools}
\small
\setlength{\tabcolsep}{6pt}
\begin{tabular}{lcccc}
\toprule
\textbf{Method} & \textbf{Electricity} & \textbf{ExchangeRate} & \textbf{Traffic} & \textbf{Weather} \\
\midrule
w/o tools
& $0.449 \pm 0.014$ & $0.205 \pm 0.003$ & $0.358 \pm 0.003$ & $0.515 \pm 0.010$ \\
w/ tools
& $0.419 \pm 0.018$ & $0.213 \pm 0.004$ & $0.335 \pm 0.003$ & $0.494 \pm 0.011$ \\
\bottomrule
\end{tabular}
\end{table}

\begin{table}[!htbp]
\centering
\caption{Mean $\pm$ standard deviation over 10 random seeds for the quality dimension design ablation study in Figure~\ref{fig:dimension_ablation}. Results report forecasting RMSE ($\downarrow$).}
\label{tab:std_dimensions}
\small
\setlength{\tabcolsep}{6pt}
\begin{tabular}{lcccc}
\toprule
\textbf{Method} & \textbf{Electricity} & \textbf{ExchangeRate} & \textbf{Traffic} & \textbf{Weather} \\
\midrule
Full
& $0.406 \pm 0.012$ & $0.211 \pm 0.006$ & $0.343 \pm 0.013$ & $0.499 \pm 0.014$ \\
w/o Data Quality
& $0.426 \pm 0.014$ & $0.214 \pm 0.005$ & $0.353 \pm 0.014$ & $0.491 \pm 0.018$ \\
w/o Rare Pattern
& $0.439 \pm 0.012$ & $0.217 \pm 0.008$ & $0.379 \pm 0.014$ & $0.509 \pm 0.019$ \\
w/o Pattern Structure
& $0.419 \pm 0.017$ & $0.220 \pm 0.005$ & $0.382 \pm 0.015$ & $0.527 \pm 0.024$ \\
\bottomrule
\end{tabular}
\end{table}

\subsection{F1 Evaluation of Perceiver Dimension Identification}
\label{app:perceiver_f1}

In addition to the Precision metric reported in Figure~\ref{fig:unit_dim_id}, we further evaluate the Perceiver's dimension identification performance using F1 score, which jointly accounts for precision and recall. As shown in Table~\ref{tab:perceiver_f1}, GRPO training consistently improves F1 scores across all four backbone models, in agreement with the precision-based results. These findings further support the effectiveness of GRPO training in identifying relevant quality dimensions.

\begin{table}[!htbp]
\centering
\caption{F1 scores ($\uparrow$) for Perceiver dimension identification with and without GRPO training across different backbone models.}
\label{tab:perceiver_f1}
\small
\setlength{\tabcolsep}{5pt}
\begin{tabular}{lcccc}
\toprule
\textbf{Method} & \textbf{Qwen3-4B} & \textbf{Gemma4-E4B} & \textbf{Phi4-mini} & \textbf{Qwen3-8B} \\
\midrule
w/o GRPO & 43.5\% & 48.2\% & 40.0\% & 48.3\% \\
w/ GRPO & \textbf{64.6\%} & \textbf{65.1\%} & \textbf{62.9\%} & \textbf{67.7\%} \\
\bottomrule
\end{tabular}
\end{table}

\subsection{Additional Baseline Comparisons for Cross-Dataset TSFM Fine-tuning}
\label{app:cross_dataset_selection}
 
We extend the cross-dataset TSFM fine-tuning experiments in Section~\ref{subsec:tsfm} by including TSRating and DataShapley as additional baselines. Following the same experimental setting, we select 50\% of the samples from the unified training pool for Timer-S1 fine-tuning and evaluate forecasting RMSE on each dataset.
As shown in Table~\ref{tab:cross_dataset_baselines}, TSQAgent achieves the lowest RMSE across all four datasets, outperforming both additional baselines and random selection.
On ExchangeRate, the improvements are relatively modest across all selection methods, potentially because Timer-S1 already achieves a low RMSE of 0.228 without fine-tuning, leaving limited room for further improvement.

\begin{table}[!htbp]
\centering
\caption{Additional baseline comparisons for cross-dataset data selection in Section~\ref{subsec:tsfm}, using Timer-S1 with a 50\% data selection budget. Results report forecasting RMSE ($\downarrow$). Best results are \textbf{bolded}.}
\label{tab:cross_dataset_baselines}
\small
\setlength{\tabcolsep}{5pt}
\begin{tabular}{lcccc}
\toprule
\textbf{Selection Method} & \textbf{Electricity} & \textbf{ExchangeRate} & \textbf{Traffic} & \textbf{Weather} \\
\midrule
Untrained   & 0.477 & 0.228 & 0.293 & 0.510 \\
Random      & 0.448 & 0.225 & 0.290 & 0.492 \\
TSRating    & 0.434 & 0.224 & 0.286 & 0.487 \\
DataShapley & 0.437 & 0.225 & 0.289 & 0.485 \\
\rowcolor{cyan!10}
\textbf{TSQAgent} & \textbf{0.425} & \textbf{0.221} & \textbf{0.283} & \textbf{0.484} \\
\bottomrule
\end{tabular}
\end{table}

\subsection{Generalization Scaling with Meta-Training Diversity}
\label{app:meta_generalization}

The cross-dataset selection experiment in Section~\ref{subsec:tsfm} meta-trains the rating model on four forecasting datasets (Electricity, ExchangeRate, Traffic, and Weather), primarily with evaluation conducted within the same forecasting task pool. We further investigate whether the learned quality representations can transfer to unseen domains and, more importantly, how such generalization scales with the diversity of the meta-training pool.

\partitle{Cross-Domain Generalization}
We first evaluate the original meta rater trained on the four forecasting datasets (\textbf{\textit{Meta Rater, 4 datasets}}) on three unseen classification datasets: MedicalImages, CBF, and BME. For each unseen dataset, we perform 5-shot adaptation and use the adapted rater to select the top 50\% of training samples for a CNN classifier. As shown in Table~\ref{tab:generalization_scaling}, the 4-dataset meta rater shows a moderate degradation compared with the \textbf{\textit{dataset-specific rater}} trained directly on each target dataset. However, it performs better than most baseline methods, suggesting it has primary generalization ability.

\partitle{Scaling Generalization through a More Diverse Task Pool}
We next investigate whether increasing the diversity of the meta-training task pool improves cross-domain generalization. We expand the meta-training pool using the Time-300B corpus~\citep{shi2025time}, selecting 23 subsets spanning 9 domains with no overlap with the three evaluation datasets. For each subset, we collect 500 pairwise quality judgments using TSQAgent, resulting in 11,500 pairwise comparisons for meta-training. We denote the resulting model as \textbf{\textit{Meta Rater, 23 datasets}}.

As shown in Table~\ref{tab:generalization_scaling}, this scaled meta rater (\textbf{\textit{Meta Rater, 23 datasets}}) achieves comparable performance to the Dataset-specific Rater and consistently outperforms Meta Rater (4 datasets) on unseen classification datasets. These results indicate a positive generalization scaling trend: increasing task diversity in meta-training can improve the generalization capability of the rating model and enable more transferable quality assessment capabilities.

\begin{table}[!htbp]
\centering
\caption{Cross-domain generalization of meta-trained rating models on three unseen classification datasets. Results report downstream classification accuracy (mean $\pm$ std.) after using each method to select the top 50\% of training samples for CNN training.}
\label{tab:generalization_scaling}
\small
\begin{tabular}{lccc}
\toprule
Method & MedicalImages & CBF & BME \\
\midrule
Random & $0.554 \pm 0.015$ & $0.595 \pm 0.061$ & $0.495 \pm 0.027$ \\
DataOob & $0.514 \pm 0.014$ & $0.663 \pm 0.016$ & $0.521 \pm 0.022$ \\
DataShapley & $0.550 \pm 0.011$ & $0.593 \pm 0.045$ & $0.564 \pm 0.028$ \\
TimeInf & $0.567 \pm 0.006$ & $0.563 \pm 0.035$ & $0.535 \pm 0.046$ \\
TSRating & $0.561 \pm 0.007$ & $0.679 \pm 0.018$ & $0.575 \pm 0.018$ \\
TSQAgent (Dataset-specific) & $0.574 \pm 0.008$ & $0.668 \pm 0.019$ & $0.580 \pm 0.021$ \\
TSQAgent (Meta Rater, 4 datasets) & $0.561 \pm 0.008$ & $0.659 \pm 0.015$ & $0.568 \pm 0.025$ \\
TSQAgent (Meta Rater, 23 datasets) & $0.571 \pm 0.009$ & $0.673 \pm 0.021$ & $0.581 \pm 0.018$ \\
\bottomrule
\end{tabular}
\end{table}

\partitle{Generalization and Computational Scalability}
As discussed in Section~\ref{subsec:cost} and Table~\ref{tab:runtime_cost}, the computational cost of TSQAgent is primarily dominated by one-time pairwise judgment annotation, while the trained meta rater can be reused with lightweight adaptation and inference. Expanding the meta-training pool from 4 datasets to 23 subsets (a 5.75$\times$ increase) requires approximately 8 hours to collect 11,500 pairwise judgments. Despite the increased annotation cost, the expanded meta rater achieves improved cross-domain generalization (Table~\ref{tab:generalization_scaling}) without increasing per-sample inference complexity. These results demonstrate the potential of scaling meta-training diversity to improve transferability while maintaining efficient deployment.

\subsection{Generalization from Synthetic Supervision to Real-World Anomalies}
\label{app:realworld_generalization}

In TSQAgent, the Perceiver is trained with GRPO using synthetic perturbations from TSQBench, where the active quality dimensions are explicitly controlled during data generation detailed in Appendix~\ref{app:benchmark}. To examine whether the quality reasoning learned from such synthetic supervision can generalize beyond synthetic perturbations, we further evaluate both the Perceiver and the complete TSQAgent framework on three real-world anomaly detection datasets: SWaT~\citep{goh2016dataset}, WADI~\citep{ahmed2017wadi}, and MSL~\citep{hundman2018detecting}. SWaT and WADI are real-world industrial sensor datasets, where anomalies reflect correlated multi-sensor changes and complex operational failures, while MSL is a spacecraft telemetry dataset containing diverse anomaly patterns arising from complex system dynamics.

\partitle{Perceiver Generalization to Real-World Quality Patterns}
We first evaluate whether the dimension selection capability learned by the GRPO-trained Perceiver transfers to real-world anomaly patterns. Since ground-truth dimension-level annotations are unavailable for these datasets, we use a specialized time series toolkit~\citep{middlehurst2024aeon} to construct weak dimension-level annotations. Specifically, the toolkit establishes normal behavior references from normal windows and identifies relevant dimensions by comparing dimension-wise deviation scores between normal--anomalous pairs, with thresholds estimated from normal--normal comparisons.

As shown in Table~\ref{tab:perceiver_realworld}, the GRPO-trained Perceiver consistently improves dimension identification over its non-GRPO counterpart across all three datasets. Precision improves from 30.4\% to 48.7\% on SWaT, from 34.8\% to 53.6\% on WADI, and from 35.0\% to 50.3\% on MSL. Similar improvements are observed in F1, suggesting that the quality dimension selection capability learned through synthetic GRPO supervision can partially transfer to real-world anomaly patterns.

\begin{table}[htbp]
\centering
\caption{Dimension identification performance of the Perceiver on real-world anomaly datasets using weak dimension-level annotations. Best results are \textbf{bolded}.}
\label{tab:perceiver_realworld}
\small
\begin{tabular}{lccc}
\toprule
& \textbf{SWaT} & \textbf{WADI} & \textbf{MSL} \\
\midrule
Precision (w/o GRPO) & 30.4\% & 34.8\% & 35.0\% \\
Precision (w/ GRPO) & \textbf{48.7\%} & \textbf{53.6\%} & \textbf{50.3\%} \\
\midrule
F1 (w/o GRPO) & 34.2\% & 39.0\% & 39.1\% \\
F1 (w/ GRPO) & \textbf{53.1\%} & \textbf{57.6\%} & \textbf{54.7\%} \\
\bottomrule
\end{tabular}
\end{table}

We also note that the Perceiver performs worse on these real-world anomaly datasets than on our TSQBench (Figure~\ref{fig:unit_dim_id}), suggesting that the current synthetic training data may not fully cover the complex and correlated quality patterns encountered in real-world scenarios. Nevertheless, the consistent improvement over the non-GRPO variant indicates that the dimension selection capability learned from synthetic perturbations remains transferable to real-world data.

\partitle{Generalization of TSQAgent to Real-World Anomalies}
We further compare TSQAgent with both time series quality assessment methods (DataOob~\citep{kwon2023data} and TimeInf~\citep{zhang2025timeinf}) and specialized anomaly detection methods (TranAD~\citep{tuli2022tranad} and USAD~\citep{audibert2020usad}) on SWaT, WADI, and MSL. DataOob and TimeInf assess sample quality from predictive utility and sample influence perspectives, respectively, while TranAD (a Transformer-based model) and USAD (an adversarial autoencoder-based model) identify anomalies by learning normal patterns and detecting deviations. TSQAgent instead learns a rater from agent-generated quality preferences over unlabeled sample pairs and identifies anomalies as samples with low quality scores.

For a fair comparison, all methods use only unlabeled sequences for training, scoring, and threshold selection, while test labels are accessed only for final F1 computation. As shown in Table~\ref{tab:realworld_anomaly}, TSQAgent achieves the highest F1 scores across all three datasets. The improvements are particularly notable on MSL and SWaT, where TSQAgent surpasses the strongest competing methods by 13.2 and 12.4 F1 points, respectively, while also achieving the best performance on WADI. These results provide further evidence that the quality rationale learned by TSQAgent from synthetic perturbations can extend beyond isolated synthetic defects to real-world data-quality failures with complex anomaly patterns.

\begin{table}[htbp]
\centering
\caption{Anomaly detection performance (F1 scores) on three real-world datasets. Best results are \textbf{bolded}.}
\label{tab:realworld_anomaly}
\small
\begin{tabular}{lccc}
\toprule
\textbf{Method} & \textbf{MSL} & \textbf{SWaT} & \textbf{WADI} \\
\midrule
DataOob & 0.575 & 0.686 & 0.671 \\
TimeInf & 0.654 & 0.473 & 0.735 \\
TranAD & 0.518 & 0.323 & 0.686 \\
USAD & 0.535 & 0.304 & 0.686 \\
\rowcolor{cyan!10}
\textbf{TSQAgent} & \textbf{0.786} & \textbf{0.810} & \textbf{0.752} \\
\bottomrule
\end{tabular}
\end{table}

The strong performance of the complete TSQAgent framework with the Perceiver-level unit test suggests that real-world quality assessment benefits from the collaborative design of the three agents. In particular, the Perceiver selects potentially relevant quality dimensions, the Inspector provides quantitative dimension-level evidence through analytical tools, and the Adjudicator integrates these intermediate results to make the final quality judgment.

\partitle{Discussion and Future Directions}
Due to the complexity of temporal patterns, obtaining reliable fine-grained quality dimension annotations for real-world time series requires substantial expert effort and annotation costs. In contrast, synthetic perturbations provide controllable and accurate ground-truth supervision at a relatively low cost, enabling a rigorous investigation of whether the Perceiver can identify relevant quality dimensions. Nevertheless, synthetic perturbations alone may not fully capture the diverse and complex quality patterns encountered in real-world scenarios. Incorporating real-world data into Perceiver training therefore represents a promising direction for improving its generalization capability. Importantly, the proposed training pipeline is not inherently limited to synthetic data and can naturally incorporate real-world quality patterns once reliable dimension-level annotations become available.

\subsection{Dimension-level Interpretability and Task-dependent Quality Assessment}
\label{app:dimension_interpretability}

Beyond overall pairwise quality assessment, TSQAgent preserves intermediate dimension-level information that supports interpretable quality judgments. We further investigate how the significance of different quality dimensions varies across downstream tasks through two case studies on anomaly detection and forecasting.

\partitle{Dimension-level Interpretability}
TSQAgent provides interpretable dimension-level evidence throughout its three-agent reasoning process. Specifically, the Perceiver identifies and ranks relevant quality dimensions for each pairwise comparison, retaining the selected dimensions for subsequent analysis. The Inspector then evaluates each selected dimension using analytical tools and generates dimension-specific evidence, including statistical measurements, detected temporal patterns, and comparative assessments. Finally, the Adjudicator integrates these intermediate results to produce the final pairwise quality judgment. By preserving the dimension selections, rankings, and supporting evidence, TSQAgent enables further inspection and interpretation of its quality assessment decisions.

\partitle{Task-dependent Dimension Significance}
To investigate whether TSQAgent captures task-dependent quality characteristics, we analyze the significance distributions of different quality dimensions on two representative datasets: MSL~\citep{hundman2018detecting} for anomaly detection and Electricity~\citep{trindade2015electricityloaddiagrams20112014} for forecasting.

On the MSL spacecraft telemetry dataset, Rare Pattern emerges as the dominant quality dimension in comparisons involving anomalous samples. As shown in Table~\ref{tab:msl_dimension}, Rare Pattern accounts for 84.95\% of the dimension significance in normal--anomalous comparisons and 72.92\% in anomalous--anomalous comparisons, compared with only 10.84\% in normal--normal comparisons. This observation is consistent with the characteristics of anomaly detection, where unusual temporal behaviors are particularly informative. Moreover, when evaluating normal--anomalous pairs under the Rare Pattern dimension, TSQAgent correctly identifies normal samples as higher-quality in 96.20\% of comparisons, demonstrating the utility of its dimension-level quality judgments.

\begin{table}[!htbp]
\centering
\caption{Distribution of quality dimension significance across different comparison types on the MSL anomaly detection dataset. The last column reports pairwise quality assessment accuracy under the Rare Pattern dimension for normal--anomalous comparisons.}
\label{tab:msl_dimension}
\small
\setlength{\tabcolsep}{3pt}
\begin{tabular}{lcccccccc}
\toprule
\textbf{Comparison} & \textbf{Trend} & \textbf{Amp.} & \textbf{Pattern} & \textbf{Freq.} & \textbf{Noise} & \textbf{Missing} & \textbf{Rare} & \textbf{Rare Acc.} \\
\midrule
Normal--Normal
& 14.61\% & 23.43\% & 20.91\% & 12.85\% & 14.36\% & 3.00\% & 10.84\% & -- \\
Normal--Anomaly
& 1.08\% & 4.30\% & 3.23\% & 0.00\% & 5.38\% & 1.08\% & \textbf{84.95\%} & \textbf{96.20\%} \\
Anomaly--Anomaly
& 2.08\% & 8.33\% & 9.38\% & 0.00\% & 4.17\% & 3.13\% & \textbf{72.92\%} & -- \\
\bottomrule
\end{tabular}
\end{table}

In contrast, the Electricity forecasting dataset exhibits a different dimension significance distribution. As shown in Table~\ref{tab:electricity_dimension}, Trend and Amplitude are the two most significant dimensions, accounting for 34.66\% and 30.16\%, respectively, or 64.82\% combined. Meanwhile, Rare Pattern contributes only 0.35\%. This distribution aligns with the characteristics of electricity forecasting, where changes in temporal trends and signal amplitudes are particularly relevant to prediction.

\begin{table}[!htbp]
\centering
\caption{Distribution of quality dimension significance on the Electricity forecasting dataset.}
\label{tab:electricity_dimension}
\small
\setlength{\tabcolsep}{5pt}
\begin{tabular}{lccccccc}
\toprule
\textbf{Dimension} & Trend & Amp. & Pattern & Freq. & Noise & Missing & Rare \\
\midrule
Significance
& \textbf{34.66\%} & \textbf{30.16\%} & 13.17\% & 11.44\% & 8.67\% & 1.56\% & 0.35\% \\
\bottomrule
\end{tabular}
\end{table}

These case studies demonstrate that TSQAgent can identify different dominant quality dimensions across downstream tasks while preserving interpretable evidence for dimension-level assessment.

\partitle{Discussion on Quality Aggregation}
In the current evaluation, we primarily aggregate dimension-level evidence into overall quality judgments to enable direct comparisons with existing data valuation methods. Nevertheless, TSQAgent retains the intermediate dimension selections, rankings, and analytical evidence, allowing its quality assessments to be interpreted at the dimension level. These capabilities also provide a basis for future task-dependent quality assessment, where different quality dimensions may be emphasized according to specific application requirements.

\raggedbottom

\section{Qualitative Case Studies}
\label{app:case_study}

We further present two qualitative case studies to provide an intuitive understanding of how GRPO-based training and tool usage improve the efficiency and reliability of TSQAgent. The first study focuses on the effect of GRPO on dimension selection behavior, while the second examines the impact of tool-augmented quantitative comparison.

\subsection{Case Study I: Effect of GRPO-based Dimension Selection.}
As shown in Figure~\ref{fig:Case Study I},  without GRPO Perceiver, the Agent tends to over-select multiple quality dimensions and produces redundant or less informative analyses. This leads to inefficient reasoning and less precise judgments.
In contrast, with GRPO Perceiver, the Agent focuses on a smaller set of truly relevant quality dimensions and provides more targeted and accurate assessments.
This case demonstrates that GRPO Perceiver improves both the precision and efficiency of quality reasoning by reducing unnecessary dimensions and emphasizing the most discriminative factors.

\subsection{Case Study II: With vs. Without Tool Augmentation.}
\begin{figure}[!htbp]
    \centering
    \begin{subfigure}[t]{0.48\linewidth}
        \centering
        \includegraphics[height=14cm,width=\linewidth]{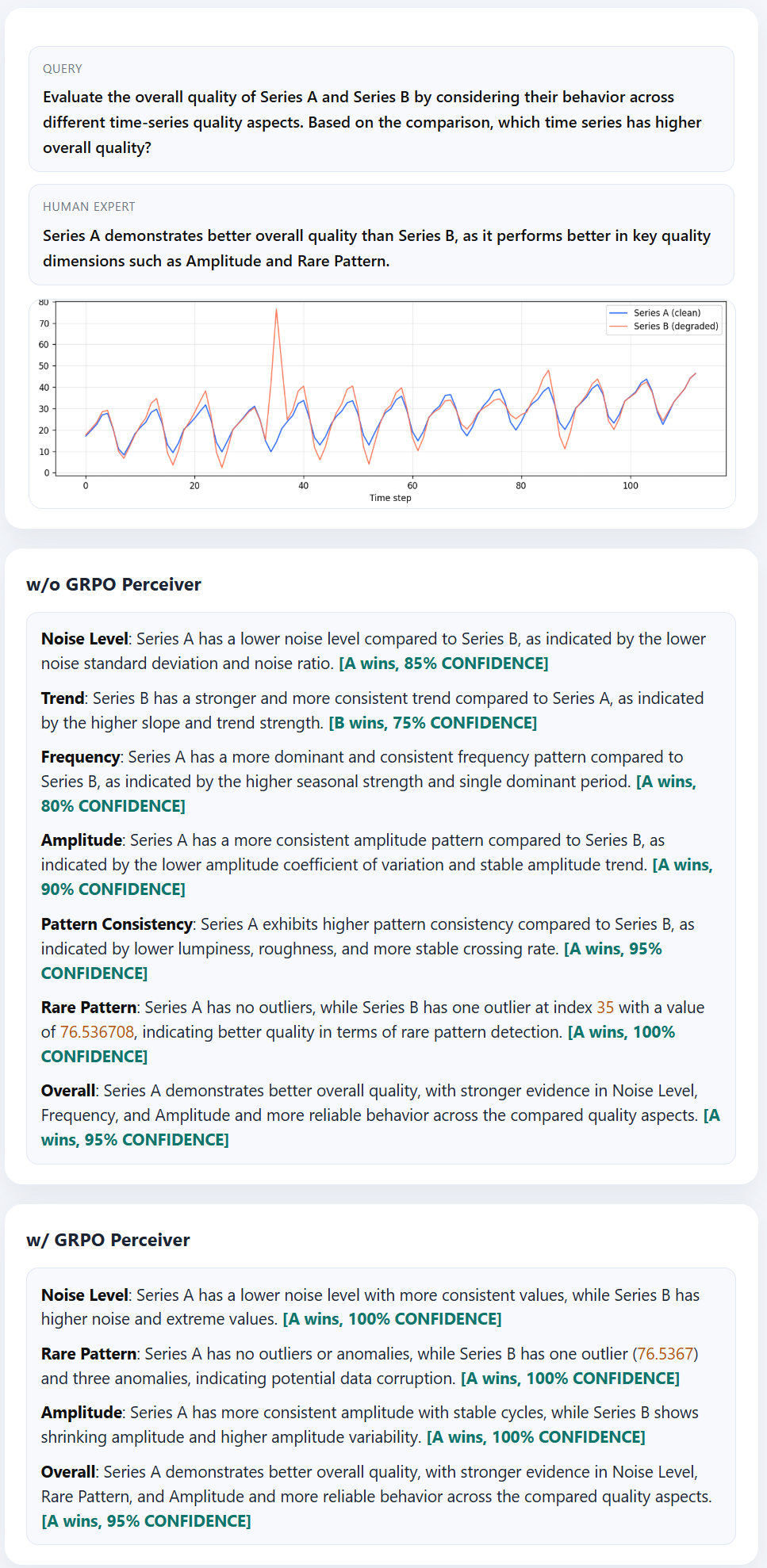}
        \caption{GRPO-based Dimension Selection}
        \label{fig:Case Study I}
    \end{subfigure}
    \hfill
    \begin{subfigure}[t]{0.48\linewidth}
        \centering
        \includegraphics[height=14cm,width=\linewidth]{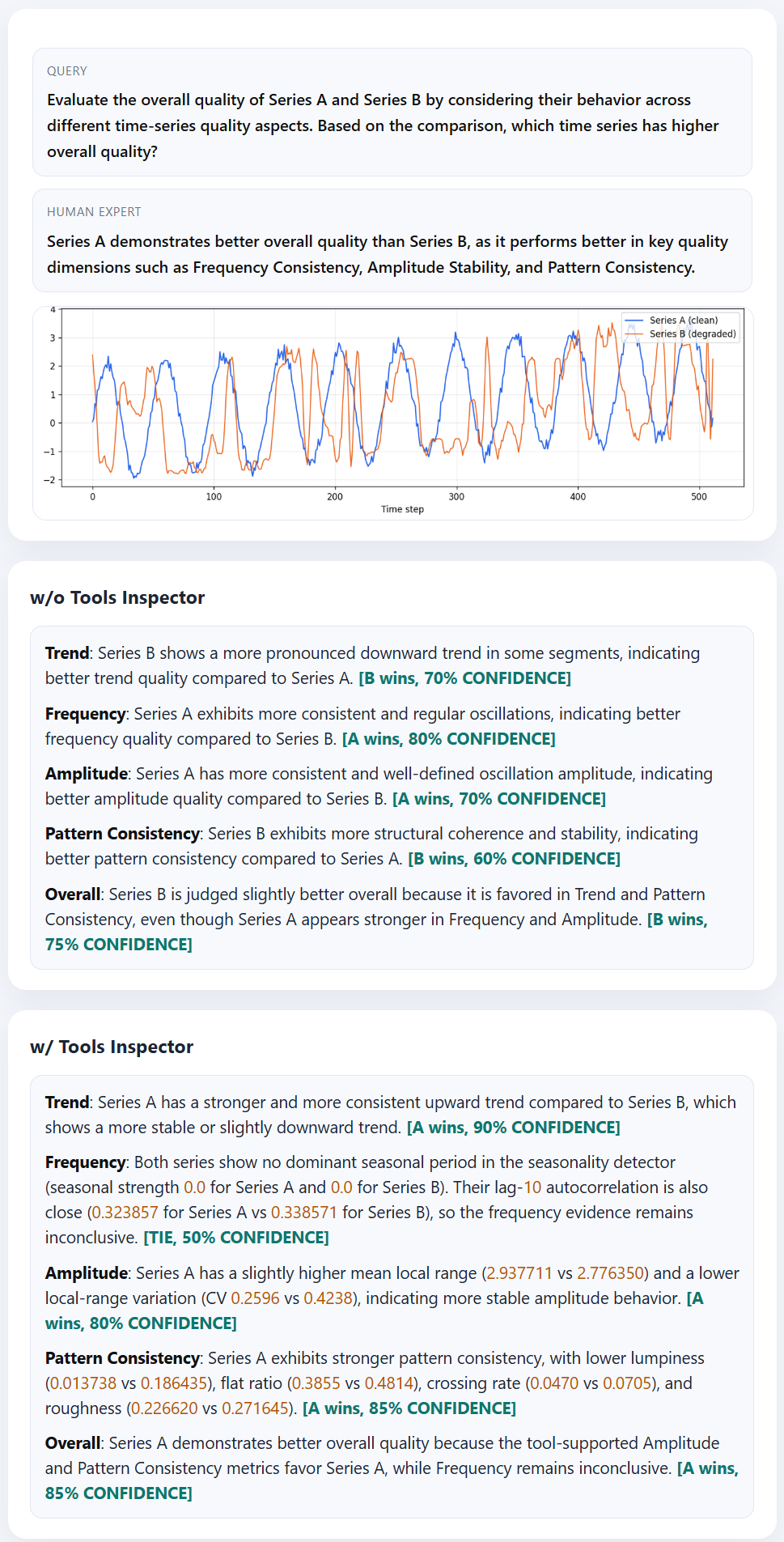}
        \caption{Tool-Augmented Quality Comparison}
        \label{fig:Case Study II}
    \end{subfigure}
    \caption{Case studies on tool augmentation and GRPO-based dimension selection. 
Each subplot shows results on a representative time series quality comparison case, including the query, human expert judgment, and agent reasoning under different configurations. 
Subplot (a) compares the Agent with and without the GRPO Perceiver, demonstrating that GRPO-based dimension selection reduces redundant quality dimensions and improves the focus of quality reasoning. 
Subplot (b) compares the Inspector with and without tool support, demonstrating that tool-augmented quantitative measurements provide more grounded evidence and improve the reliability of the final quality judgment.}
    \label{fig:case_study}
\end{figure}

As shown in Figure~\ref{fig:Case Study II},  when tool usage is disabled, the Inspector mainly relies on subjective reasoning, leading to inconsistent evaluations across quality aspects. It overemphasizes trend and pattern consistency, resulting in an incorrect overall judgment favoring Series B. 
When tool support is enabled, the Inspector incorporates quantitative measurements from external tools, leading to more grounded and consistent assessments. As a result, it correctly identifies Series A as having higher overall quality.
This case shows that tool-based evidence improves the reliability and precision of the Inspector by reducing subjective bias.

\subsection{Extended Case Studies}

We provide extended qualitative case studies, as shown in Figure~\ref{fig:case_study_all}. These cases further demonstrate the effectiveness of our framework in identifying relevant quality dimensions and producing consistent judgments under diverse time series degradation scenarios.

\begin{figure}[!htbp]
    \centering

    \begin{subfigure}[t]{0.48\textwidth}
        \centering
        \includegraphics[width=\linewidth,height=9cm]{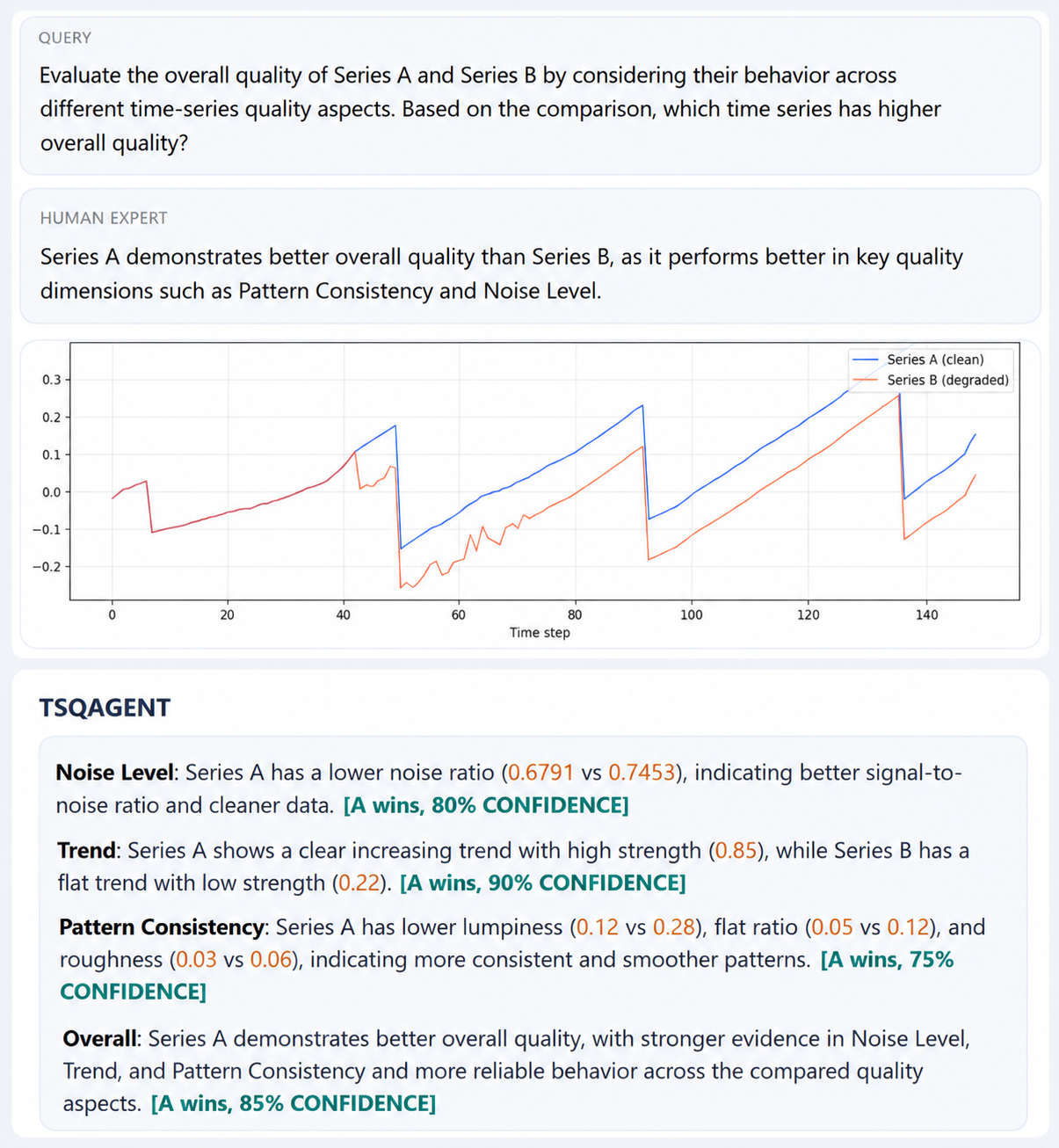}
        \caption{Case involving pattern consistency and noise level}
        \label{fig:case1}
    \end{subfigure}
    \hfill
    \begin{subfigure}[t]{0.48\textwidth}
        \centering
        \includegraphics[width=\linewidth,height=9cm]{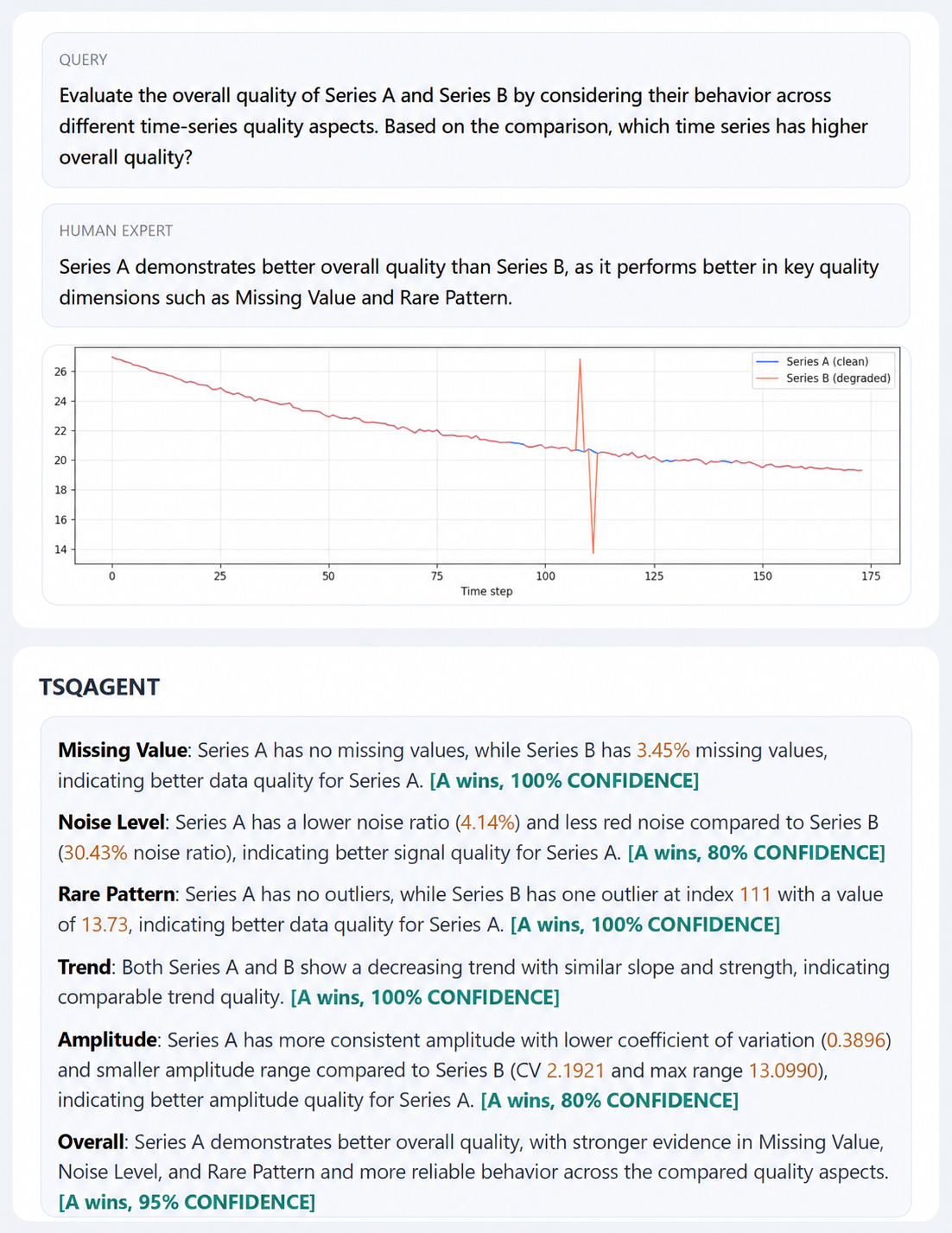}
        \caption{Case involving missing value and rare pattern}
        \label{fig:case2}
    \end{subfigure}
\end{figure}

\begin{figure}[!htbp]
    \ContinuedFloat
    \centering
    \setcounter{subfigure}{2}

    \begin{subfigure}[t]{0.48\textwidth}
        \centering
        \includegraphics[width=\linewidth,height=9cm]{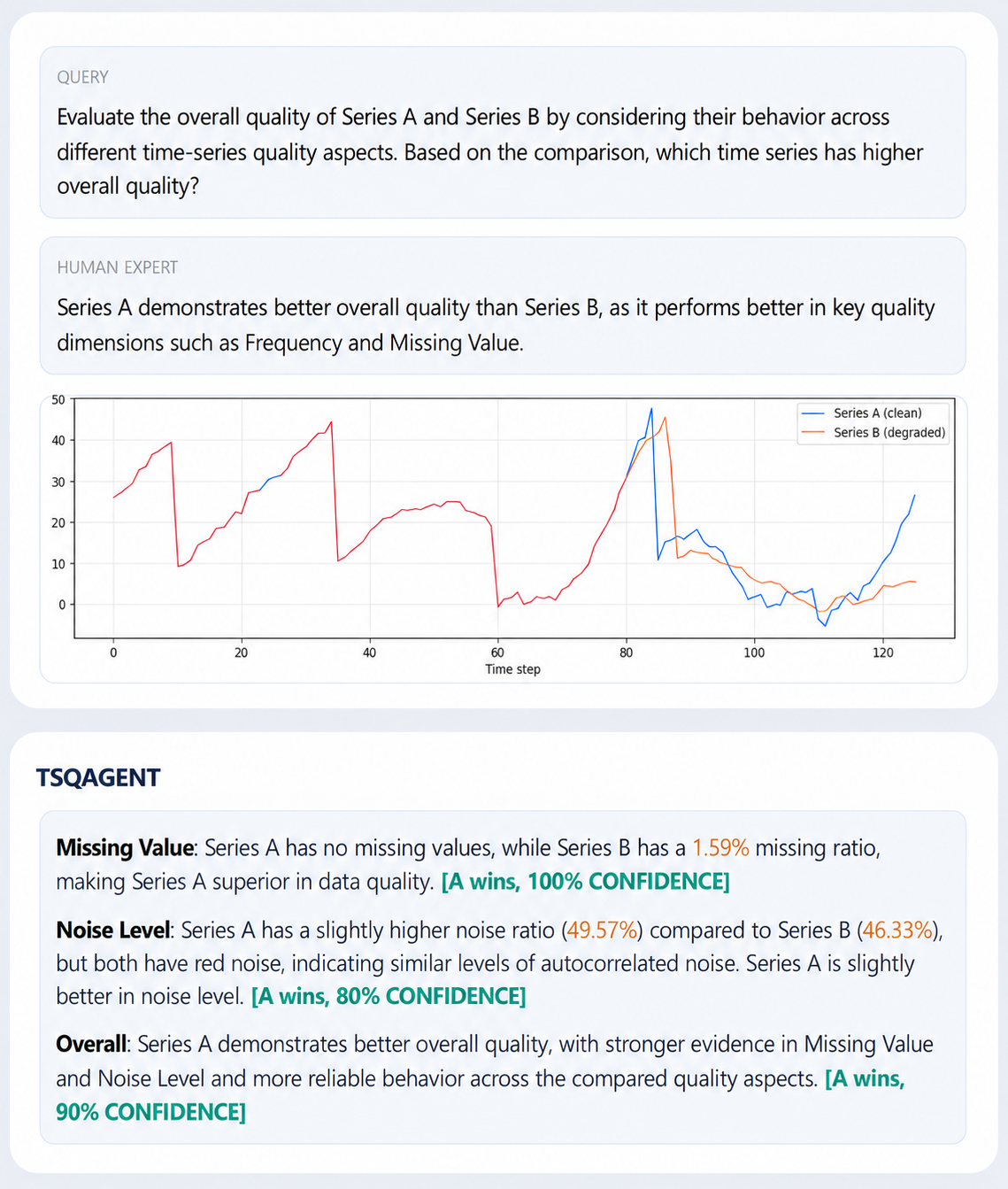}
        \caption{Case involving frequency and missing value}
        \label{fig:case3}
    \end{subfigure}
    \hfill
    \begin{subfigure}[t]{0.48\textwidth}
        \centering
        \includegraphics[width=\linewidth,height=9cm]{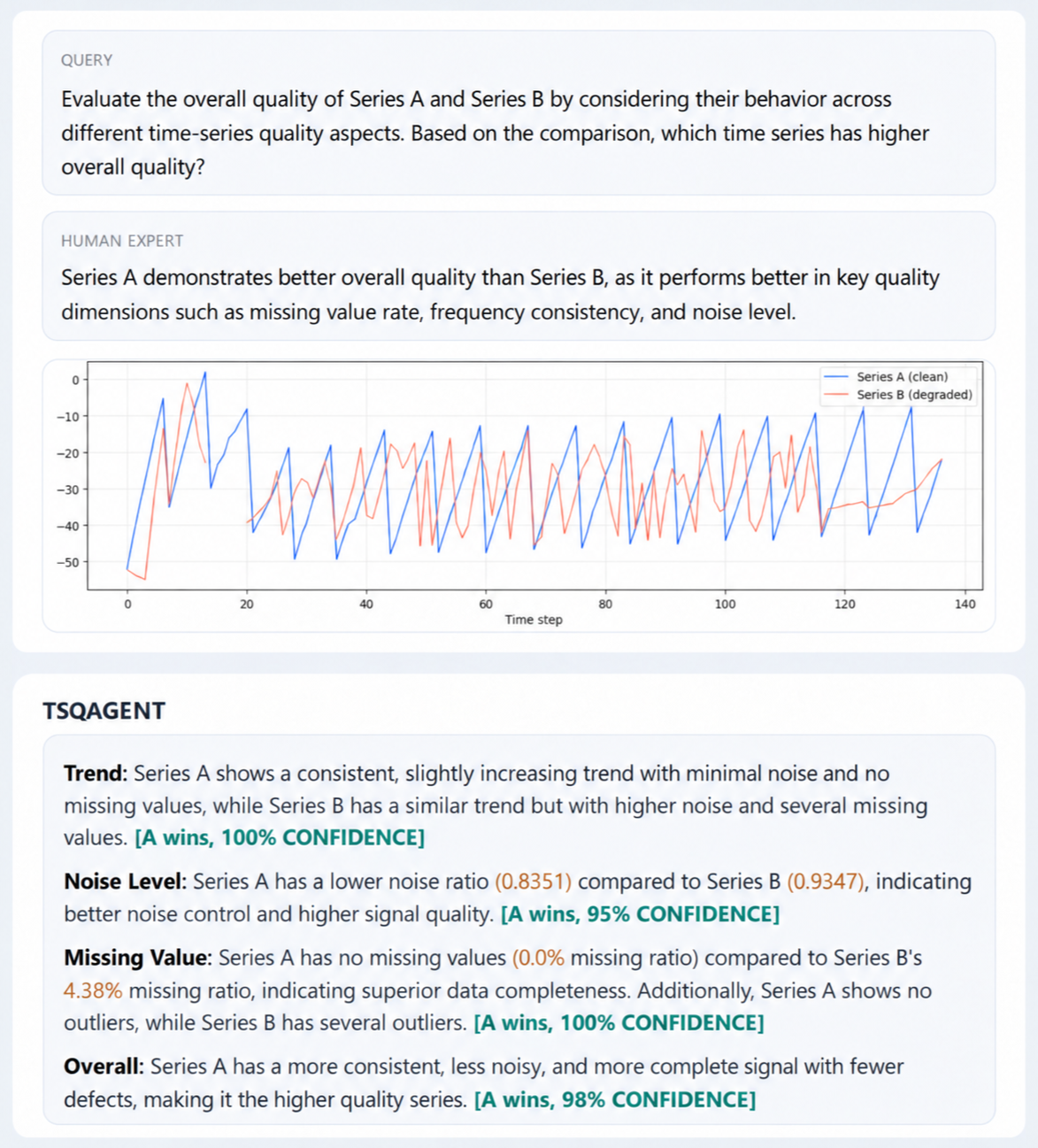}
        \caption{Case involving missing value, pattern consistency and noise level}
        \label{fig:case4}
    \end{subfigure}
\end{figure}

\begin{figure}[!htbp]
    \ContinuedFloat
    \centering
    \setcounter{subfigure}{4}

    \begin{subfigure}[t]{0.48\textwidth}
        \centering
        \includegraphics[width=\linewidth,height=9cm]{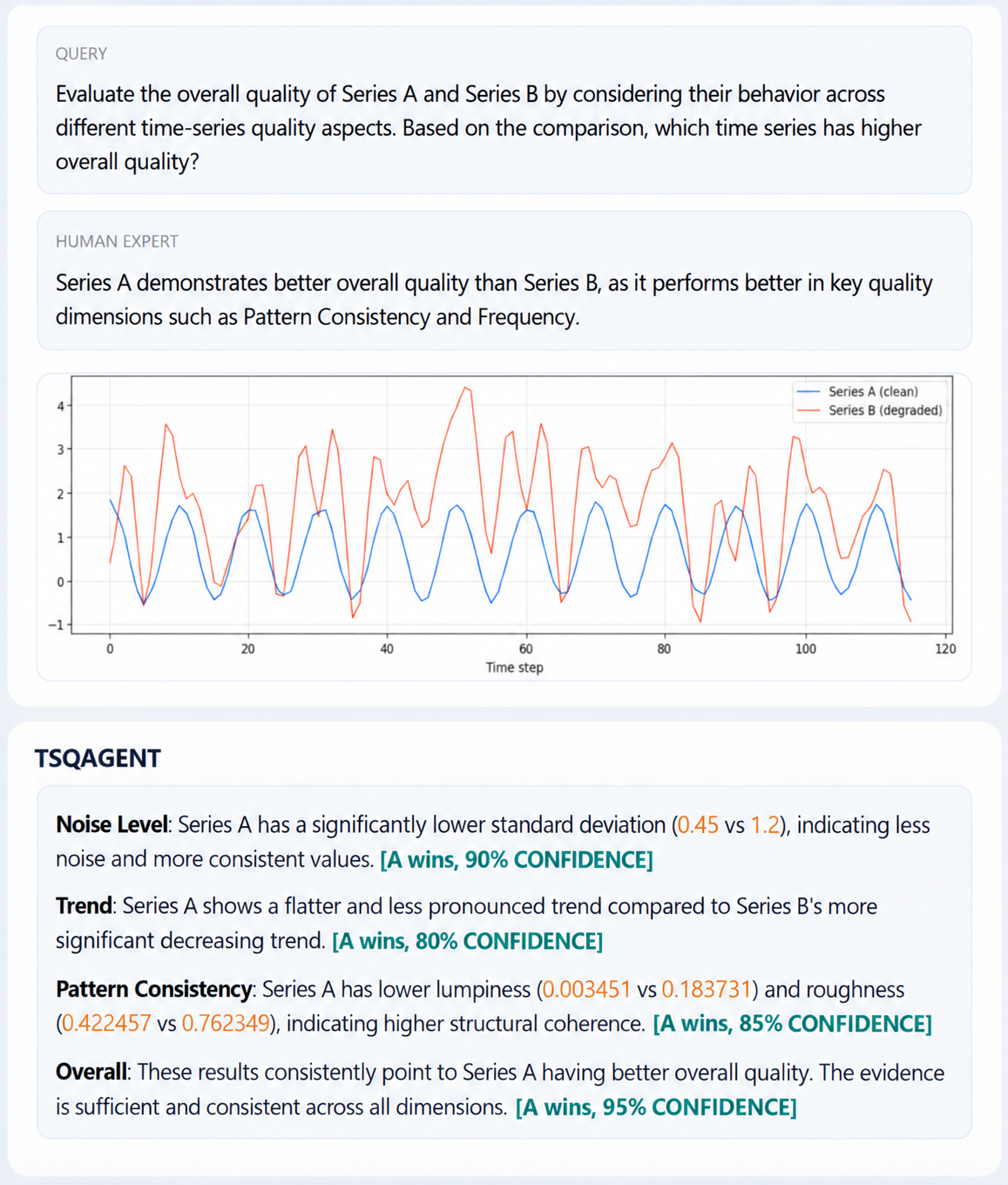}
        \caption{Case involving pattern consistency and frequency}
        \label{fig:case5}
    \end{subfigure}
    \hfill
    \begin{subfigure}[t]{0.48\textwidth}
        \centering
        \includegraphics[width=\linewidth,height=9cm]{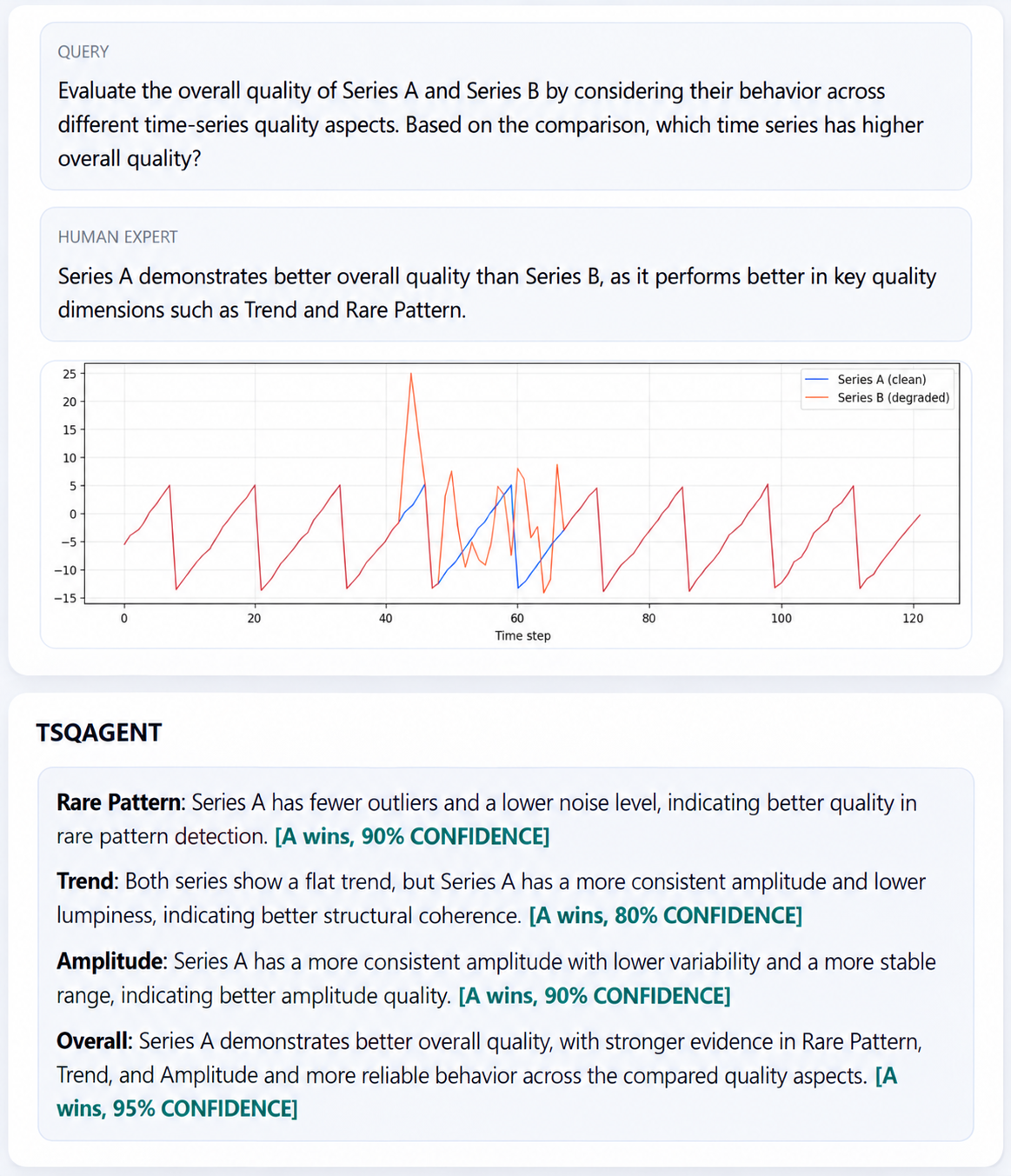}
        \caption{Case involving trend and rare pattern}
        \label{fig:case6}
    \end{subfigure}

    \caption{Extended qualitative case studies under six representative quality comparison scenarios. Each subfigure illustrates the reasoning process and final judgment produced by TSQAgent.}
    \label{fig:case_study_all}
\end{figure}

\section{Limitations}
\label{app:limitations}

Despite the effectiveness of TSQAgent, several limitations remain that warrant further investigation.
First, TSQBench relies on controlled synthetic defect injection to construct quality differences between time series. Although this design enables systematic supervision and evaluation across predefined quality dimensions, synthetic perturbations may not fully reflect the complexity of real-world data quality issues, where multiple defects can interact and their manifestations may vary across domains. Consequently, the Perceiver's learned dimension selection policy may be influenced by the distributions and degradation patterns represented in TSQBench, potentially limiting its generalization to unseen or domain-specific quality issues. Future work could incorporate more diverse real-world quality annotations and degradation patterns to reduce this dependence on synthetic supervision.

Second, TSQAgent relies on a predefined set of seven quality dimensions, which may not fully capture the diverse quality characteristics encountered in different application domains. In particular, certain domain-specific quality issues may fall outside the current dimension set, potentially limiting the framework's applicability to more specialized scenarios. Adapting the framework to such scenarios may therefore require manually identifying and incorporating additional quality dimensions. Future work could explore more flexible and extensible quality dimension definitions to accommodate diverse data characteristics and application requirements.

Third, the tool-augmented Inspector introduces additional inference costs through iterative LLM reasoning and analytical tool invocation. These costs can increase with the number of selected quality dimensions, tool calls, and reflection steps, particularly when assessing large collections of time series. Although bounded reasoning steps and caching help control computational overhead, applying the full agentic pipeline to every candidate pair may still be expensive in large-scale data selection scenarios. Future work could explore more efficient tool selection, early stopping, and cost-aware agent coordination~\citep{lin2026squidagent} to reduce inference overhead.

Finally, the scalability of TSQAgent to substantially larger data pools remains to be systematically evaluated. Although the rating model supports sample-level quality scoring and ranking, obtaining sufficient pairwise judgments for large-scale data selection may introduce additional computational costs and sampling challenges. Future work could investigate more efficient comparison sampling and ranking strategies, while further evaluating the framework across a broader range of downstream tasks and application domains.

\section{Broader Social Impacts}
\label{app:broader_impacts}

This work focuses on time series quality assessment and data selection using large language models and agent-based reasoning. The proposed framework may have positive societal impacts by improving the efficiency and reliability of time series modeling pipelines, which are widely used in real-world applications such as energy systems, finance, healthcare, and environmental monitoring. By enabling more effective identification of high-quality data, the method can contribute to more accurate forecasting and decision-making in these domains.

At the same time, potential negative impacts should also be considered. Since the proposed method involves automated evaluation and selection of training data, improper use may amplify biases present in the data or lead to suboptimal model behavior if deployed without proper validation. In high-stakes applications, reliance on automated quality assessment without human oversight may introduce risks, especially when data distributions are shifted or noisy. Therefore, we recommend that the system be used with human-in-the-loop verification in sensitive scenarios.


\end{document}